\documentclass[preprint,12pt,authoryear]{elsarticle}

\usepackage{hyperref}
\usepackage{amsmath}
\usepackage{amssymb}
\usepackage{adjustbox}
\usepackage{lscape}
\usepackage{comment}
\usepackage{xcolor}

\journal{Nuclear Physics B}

\begin{document}

\begin{frontmatter}



\begin{comment}
1. Advances in Legal Question Answering: A Comprehensive Survey
2. Exploring the State of the Art in Legal QA Systems    
3. Deep Learning for Legal QA: A Review of Benchmark Datasets
4. Question-Answering in the Legal Domain: An Overview
5. A Review of Legal QA Systems and Datasets: Current Trends and Future Directions
6. Legal QA: A Review of Benchmark Datasets and Progress in the Field"
\end{comment}
\title{Exploring the State of the Art in Legal QA Systems }


\author[inst1,inst2]{Abdelrahman Abdallah\corref{equal_contrib}}
\author[inst1]{Bhawna Piryani\corref{equal_contrib}}
\author[inst1]{Adam Jatowt}

\affiliation[inst1]{organization={Department of Computer Science \& DiSC, University of Innsbruck},%Department and Organization
            addressline={Innrain 52}, 
            city={Innsbruck},
            postcode={6020}, 
            country={Austria}}
\affiliation[inst2]{organization={Information Technology Department},%Department and Organization
            addressline={Assiut University}, 
            city={Assiut},
            postcode={71515}, 
            state={Assiut},
            country={Egypt}}
\cortext[equal_contrib]{Equal contribution.}
\cortext[corresponding_author]{Corresponding author: Abdelrahman Abdallah. Email: abdelrahman.abdallah@uibk.ac.at}

\begin{abstract}
Answering questions related to the legal domain is a complex task, primarily due to the intricate nature and diverse range of legal document systems. Providing an accurate answer to a legal query typically necessitates specialized knowledge in the relevant domain, which makes this task more challenging, even for human experts.  Question answering (QA) systems are designed to generate answers to questions asked in natural languages. QA uses natural language processing to understand questions and search through information to find relevant answers. At this time, there is a lack of surveys that discuss legal question answering. To address this problem, we provide a comprehensive survey that reviews 14 benchmark datasets for question-answering in the legal field as well as presents a comprehensive review of the state-of-the-art Legal Question Answering deep learning models.  We cover the different architectures and techniques used in these studies and discuss the performance and limitations of these models. Moreover, we have established a public GitHub repository that contains a collection of resources, including the most recent articles related to Legal Question Answering, open datasets used in the surveyed studies, and the source code for implementing the reviewed deep learning models\footnote{The repository is available at: \url{https://github.com/abdoelsayed2016/Legal-Question-Answering-Review}.}. The key findings of our survey highlight the effectiveness of deep learning models in addressing the challenges of legal question answering and provide insights into their performance and limitations in the legal domain.
\end{abstract}
\end{frontmatter}

\section{Introduction}
\label{sec:Introduction}

QA \citep{choi2018quac,allam2012question} is a kind of artificial intelligence (AI) task intended to provide answers to queries in a natural language like humans do. NLP (Natural language processing) methods are generally used in QA systems to grasp the meaning of the question and then apply various techniques such as machine learning and information retrieval to locate the most suitable answers from a large pool of data. Deep learning-based QA is a trending field of AI \citep{weston2015towards}  that employs deep learning techniques to build QA systems. Deep learning is a form of machine learning where neural networks with multiple layers are used to comprehend complex patterns in data. In the domain of QA, deep learning methods can be utilized to enhance the system's capability to understand the meaning of a question and locate the most appropriate answer from a large pool of data.

Deep learning has become popular in the recent years and have been used to build state-of-the-art QA systems that provide answers with high accuracy for a wide range of questions. Some examples of question-answering systems that use deep learning include Generative Pretrained Transformer 3 (GPT-3) \citep{brown2020language} and Google's BERT \citep{devlin2018bert,qu2019bert,wang2019multi,kassner-schutze-2020-bert}. Deep learning has many significant advantages for question-answering tasks. One of the main benefits of deep learning is that it allows QA systems to handle complex and unstructured data \citep{komeili2021internet,khandelwal20generalization}, such as natural language text, more effectively than other machine learning techniques. This is because deep learning models can learn to extract and interpret the underlying meaning of a question and its context rather than just relying on pre-defined rules, statistical patterns, or hand-crafted features. Another key benefit of deep learning for QA is that it allows for end-to-end learning, where the entire system, from input to output, is trained together. This can improve the QA system's overall performance and make it easier to train and maintain models. Finally, deep learning \citep{perez2020unsupervised,lewis2019unsupervised,lin2017structured} also enables the use of large-scale, unsupervised learning where the model can learn from vast amounts of unlabeled data. This can be particularly useful for QA systems, as it allows them to learn from various sources and improve their performance over time. To summarize, the use of deep learning in question answering has helped make QA systems more accurate and effective and has opened up new possibilities for using AI to answer a wide range of questions. When using deep learning to answer questions, it is important to use neural network architectures specifically designed for QA tasks.

These architectures typically consist of multiple layers of interconnected nodes, which are trained to process the input data and generate a response. Information Retrieval (IR) \citep{yang2019hybrid,34774953531722,abdallah2023generator} approaches can be used to find the most relevant documents or passages from a corpus of text containing the required information to solve a given question. Typically, the procedure comprises assessing the question to identify relevant keywords, followed by a search for relevant documents or passages in the collection using those keywords.

For example, one common architecture for QA is the encoder-decoder model \citep{chen2020multimodal,golub2016character}, where the input question is first passed through an "encoder" network that converts it into a compact representation. This representation is then passed through a "decoder" network module that generates the answer. The encoder and decoder networks can be trained together using large amounts of labeled data, where the correct answers are provided for a given set of questions. Another popular architecture for QA is the transformer model, which uses self-attention mechanisms \citep{he2016character,nie2017attention,abdallah2020attention} to allow the model to focus on different parts of the input data at different times. This enables the model better to capture the meaning and context of the question and generate more accurate answers. Overall, using these specialized neural network architectures has been the key to the success of deep learning for question-answering and has enabled the development of highly effective QA systems. While deep learning has made significant progress in question answering, there are still many challenges \citep{reddy2019coqa,talmor2018commonsenseqa,ezzeldin2012survey} that need to be addressed in order to make QA systems even more effective and useful.

\begin{figure}[h!]
    \centering
   
        \includegraphics[width=0.7\textwidth]{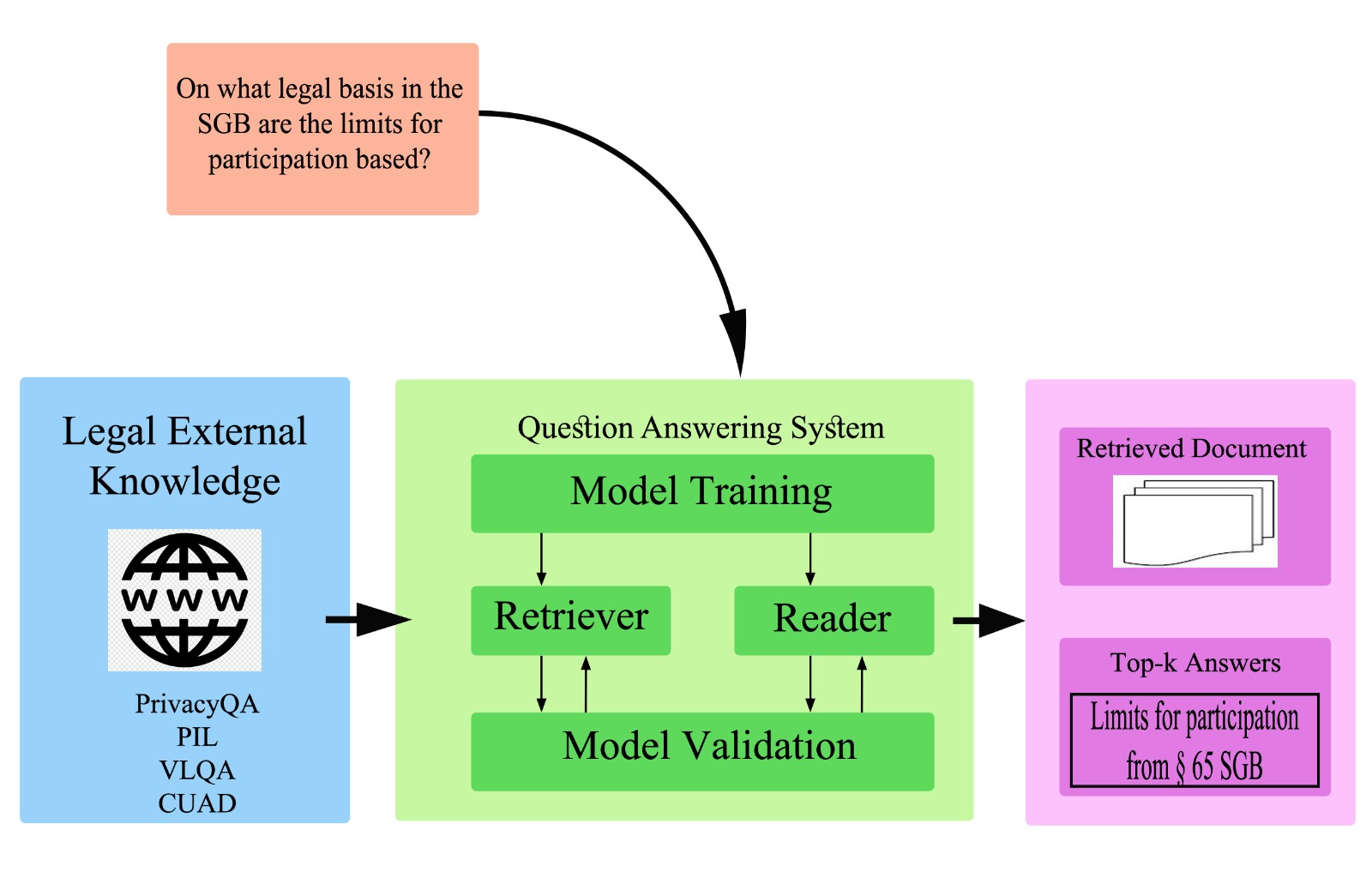}

\caption{Overview of QA Research Framework combining methods, datasets, and models.} 
\label{fig:pipline}
\end{figure}

To provide a comprehensive overview of typical QA steps, we present Figure \ref{fig:pipline}, which illustrates the QA Research Framework. This figure combines various QA methods, datasets, and models, highlighting the interplay between these components and their significance in the field.
\subsection{Overview of Legal QA }

Legal question answering (LQA) \citep{fawei2019semi,morimoto2017legal} is the process of providing answers to legal questions. Usually, a lawyer or another legal professional with expertise and knowledge in the relevant area of law does this. Legal question answering may involve various actions, including researching the existing law, interpreting legal statues and regulations, and applying legal principles and precedents to specific factual situations. LQA aims to provide accurate and reliable information and advice on legal matters to help individuals and businesses navigate the legal system and resolve legal issues. Legal question answering using deep learning \citep{do2017legal,kim2015applying,kim2015convolutional} is a kind of natural language processing (NLP) task that uses machine learning algorithms to provide answers to legal questions. This approach uses deep learning, which is a subset of machine learning that involves training neural network models on large amounts of data to learn complex patterns and relationships.

In the context of legal question answering, deep learning algorithms can be trained on a large dataset of legal questions and answers to learn how to generate answers to new legal questions automatically. The algorithms can analyze the input question, identify the relevant legal concepts and issues, and generate an appropriate response based on the learned patterns and relationships in the data.

The legal profession is intricate and dynamic, making it an ideal candidate for QA implementation, yet one that poses also many challenges. By automating the process of looking through massive volumes of data, these technologies can assist professionals like lawyers in discovering the required information more quickly. One of the most important applications of quality assurance systems in law is legal research, where LQA technologies can be used to obtain pertinent case law and statutes quickly and to discover prospective precedents and issues of conflict. Moreover, QA systems can aid with contract review, legal writing, and other legal tasks.

\begin{figure}[h!]
    \centering
   % \begin{subfigure}{\columnwidth}
        \includegraphics[width=0.7\textwidth]{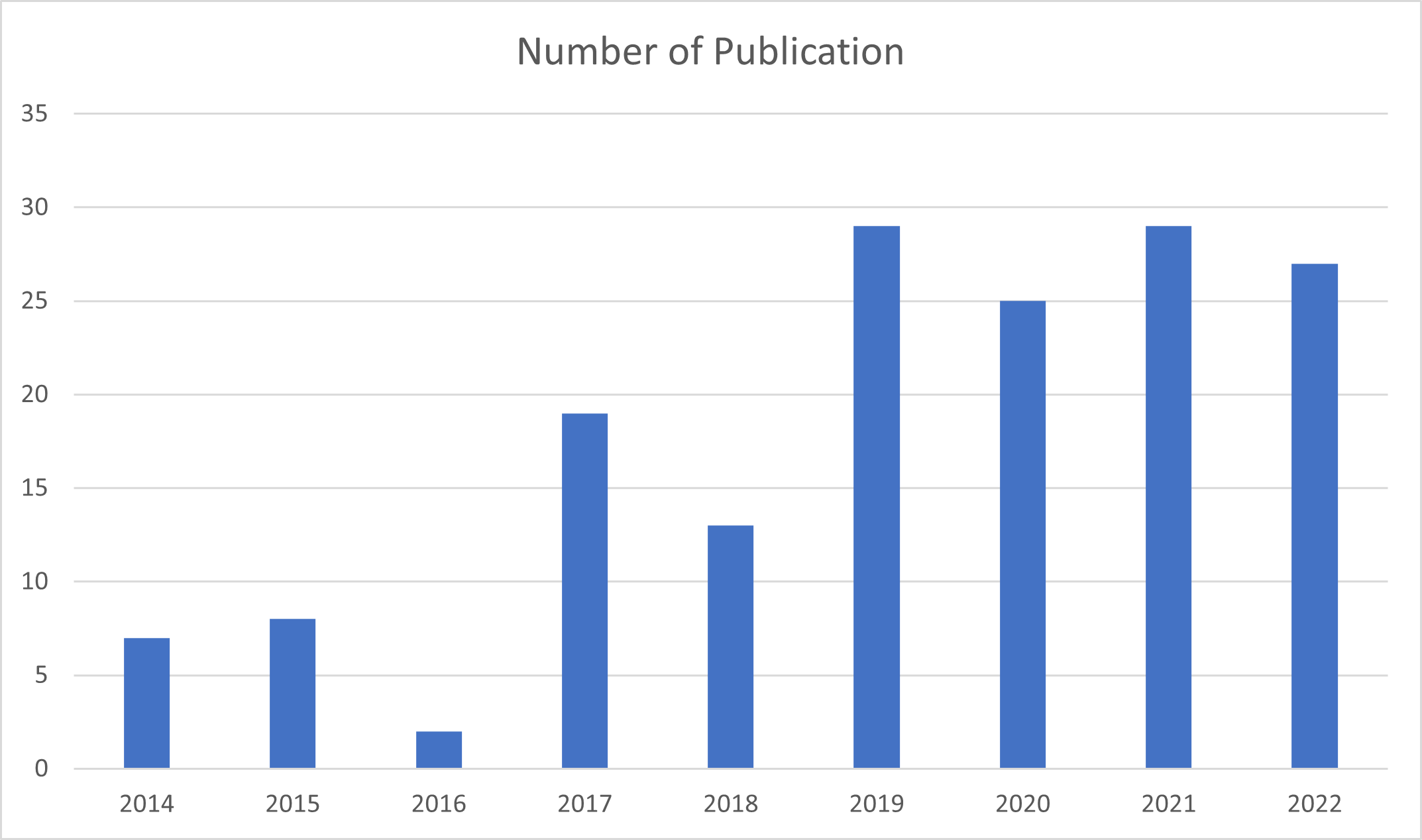}

        %\caption{}
        %\label{fig:example4}
    %\end{subfigure}
\caption{A growing trend of papers dedicated to Question Answering in the field of Law. The graph was generated by reviewing yearly publications  from 2014 to 2022 based on the data obtained from Scopus.} 
\label{fig:Chart about Table analysis publications}
\end{figure}

Figure \ref{fig:Chart about Table analysis publications} depicts the number of articles released each year that investigate deep learning strategies for different LQA challenges left. We obtained this figure from Scopus, a comprehensive bibliographic database. The search was conducted using specific keywords, including \textit{legal}, \textit{question answering}, and \textit{deep learning}. One can observe that the number of publications has been steadily growing in recent years. From 2014 to 2016, only around 17 relevant publications were published per year. Since 2017, the number of papers has significantly increased because many researchers have tried diverse deep-learning models for QA in many application fields. There are around 19 relevant articles published in 2019, which is a significant quantity. Because of the diversity of applications and the depth of challenges, there is an urgent need for an overview of present works that investigate deep learning approaches in the fast-expanding area of QA for the following reasons. It may show the commonalities, contrasts, and broad frameworks of using deep learning models to solve QA issues. This allows for the exchange of approaches and ideas across research challenges in many application sectors.

%Figure \ref{fig:Chart about Table analysis publications} depicts the number of articles released each year that investigate deep learning strategies for different LQA challenges left. One can observe that the paper number has been steadily growing in recent years. From 2014 to 2016, only around 17 relevant publications were published per year. Since 2017, the number of papers has significantly increased because many researchers have tried diverse deep-learning models for QA in many application fields. There will be around 19 relevant articles published in 2019, which is a significant quantity. Because of the diversity of applications and the depth of challenges, there is an urgent need for an overview of present works that investigate deep learning approaches in the fast-expanding area of QA for the following reasons. It may show the commonalities, contrasts, and broad frameworks of using deep learning models to solve QA issues. This allows for the exchange of approaches and ideas across research challenges in many application sectors.

\subsection{Our contributions}

In this paper, we provide a comprehensive review of recent research on legal question-answering systems. We made sure that the survey is written in accessible way as it is meant for both computer science scholars as well as legal researchers/practitioners. Our review highlights the key contributions of these studies, which include the development of new taxonomies for legal QA systems, the use of advanced NLP techniques such as deep learning and semantic analysis, and the incorporation of abundant resources such as legal dictionaries and knowledge bases. Additionally, we discuss the various challenges that legal QA systems still face and potential directions for future research in this field. Other contributions that we discuss include the use of FrameNet, ensemble models, Reinforcement Learning, multi-choice question-answering systems, legal information retrieval, the use of different languages like Japanese, Korean, Vietnamese, and Arabic, and techniques like dependency parsing, lemmatization, and word embedding. Our key contributions include the following:
\begin{enumerate}
    \item We provide a taxonomy for legal question-answering systems which categorizes legal question-answering systems based on the type of question and answer, the type of knowledge source they use, and the technique they employ and provide a clear and organized overview of  field and allow for a better understanding of the various approaches used in legal question answering, by classify system according to the domain, question type, and approach.

    \item We provide a comprehensive review of the recent development in legal question-answering system, highlighting their key contribution and similarities. We are discussing a wide range of studies, from an early study that focuses on answering yes/no questions on legal bar examination to a recent study that employs deep learning techniques for more challenging questions. Our review provides an in-depth understanding of the state-of-the-art in legal question answering and highlights the key advancement in the field.
    
    \item We list available datasets for readers to refer to, including notable studies and their key contributions. The extensive list of studies discussed in this paper provides a starting point for further research, and the taxonomy introduced in this paper can serve as a guide for the design of new legal question-answering systems.

\end{enumerate}

The remainder of the paper is structured as follows: We discuss QA challenges and ethical and legal aspects of legal Q\&A, and compare and contrast them in the subsequent subsections. Section 2 presents a review of related works in the field of legal question answering. Section 3 summarizes and explores classical and modern machine learning for question answering. Section 4 outlines the datasets and availability of source codes utilized in reviewed studies and offers an overview of resources available for replication and comparison of LQA. In Section 5, we assess the performance of LQA models, emphasizing their strengths and limitations. Lastly, in Section 6, we draw conclusions and suggest future research directions in the field of legal question answering.

\subsection{QA challenges}
One of the main challenges in generic QA is the inherent complexity of natural language. Human language is highly nuanced and contextual and often uses multiple meanings and ambiguities. This can make it difficult for many QA systems to understand a question's meaning accurately and generate the correct answer.

Another challenge is the lack of high-quality, labeled training data \citep{nguyen2019overcoming}. QA systems require large amounts of data to learn from, but it can be difficult and time-consuming to create and annotate such data manually. This can limit the performance of QA systems, especially when they are trained on small or noisy datasets. Large amounts of labeled data are needed to train a quality assurance (QA) model. Providing the correct answer to a given question is used to label the data. This process is frequently labor-intensive and time-consuming. In addition, high-quality training data should be diverse and representative of the question types that the QA system will be expected to answer. However, it is frequently challenging to produce or find diverse and representative high-quality training data. There are also issues with the representation of the data. For example, a QA system that is trained only in a specific domain, like Law or Medicine, may not perform well when it's asked questions from different domains or general domains.

However, there is also the challenge of ensuring that QA systems are trustworthy and provide reliable answers. As AI systems become more widely used, it is important to ensure that they are transparent and accountable and that they do not perpetuate biases or misinformation \citep{cadene2019rubi,pal2012exploring}. High-quality, labeled training data is essential for training QA models to comprehend and respond to questions accurately. If the training data are unrepresentative or of poor quality, the performance of the system may suffer. Typically, reliable and Trustworthy QA systems are developed using strong models that can generalize well to new questions. This indicates that they are able to respond accurately to inquiries that they have never seen before.

%%%%%%%%%%%%%%%%%%%%%%%%%%%%%%%%%%%%%
Finally, there are also some general challenges to using QA systems in domain-specific fields such as law and medicine. One major challenge is the complexity and ever-changing nature of the information in these fields, which can make it difficult for QA systems to stay up-to-date. Additionally, there may be ethical and legal considerations to especially take into account when using QA systems in these fields, such as concerns about data privacy and patient confidentiality.
%%%%%%%%%%%%%%%%%%%%%%%%%%%%%%%%%%%%%
\subsection{Generic VS Legal Question Answering Systems}%Comparing and Contrasting}
The key differences between generic QA systems, and legal QA systems can be summarized as follows:
\begin{itemize}

    \item Quality and Reliability: The output of legal QA systems must have a high level of quality and reliability as their output can have a direct impact on the outcome of a case.
    \item Domain expertise: Generic QA systems have a broad understanding of various topics, while legal QA systems have a specialized understanding of their respective fields.
    \item Data: The training and testing data for these systems require specialized sets of data that are not found in a generic QA systems training set.
    \item Updating a legal QA system: laws and regulations can change frequently and can be complex, and new laws and regulations may be needed to be added to the training data or underlying dataset for developing QA models.
    \item Data privacy and security: Legal QA systems deal with sensitive information and need to be designed with strong security measures to protect client privacy and comply with regulations.

\end{itemize}
For a more detailed comparison between legal Q\&A and general Q\&A, we refer readers to Table \ref{tab:my_label_comparsion}, which provides an overview of key differences and similarities in various aspects of legal and general Q\&A
\begin{table}[ht]
    \centering
    \begin{adjustbox}{width=1\textwidth}
    \begin{tabular}{|c|c|c|}
    \hline
    \textbf{Aspect} & \textbf{Legal Q\&A} & \textbf{General Q\&A} \\ \hline
    Knowledge required & \begin{tabular}[c]{@{}c@{}} Specialized knowledge of \\laws and regulations\end{tabular} & \begin{tabular}[c]{@{}c@{}} Can vary depending \\on topic \end{tabular} \\ \hline
    %\begin{tabular}[c]{@{}c@{}}Implications of inaccurate answers \end{tabular} & Serious & Less serious \\ \hline
    %\begin{tabular}[c]{@{}c@{}}Research and \\consultation required \end{tabular} & \begin{tabular}[c]{@{}c@{}}Often requires additional \\ research and consultation with \\legal experts \end{tabular}& \begin{tabular}[c]{@{}c@{}} May not require \\additional research or \\consultation with experts \end{tabular} \\ \hline
    Complexity &  \begin{tabular}[c]{@{}c@{}}Often complex \\ and nuanced \end{tabular} & \begin{tabular}[c]{@{}c@{}} Can vary but may \\ be more straightforward \\ than legal Q\&A \end{tabular}\\ \hline
    Consequences & \begin{tabular}[c]{@{}c@{}}Can have serious \\ legal and financial \\consequences\end{tabular} & \begin{tabular}[c]{@{}c@{}}Usually less severe\\ consequences\end{tabular} \\ \hline
    Sources & \begin{tabular}[c]{@{}c@{}} Legal texts, \\case law, statutes,\\ and regulations \end{tabular}& \begin{tabular}[c]{@{}c@{}}Can vary depending \\ on the topic, and may \\include expert opinions, \\published research, \\and common knowledge \end{tabular}\\ \hline
    %Expertise & \begin{tabular}[c]{@{}c@{}} Requires knowledge of\\ legal principles and \\procedures\end{tabular} &\begin{tabular}[c]{@{}c@{}}  Can vary depending on \\the topic, may include\\ knowledge of a specific\\ subject or industry \end{tabular}  \\ \hline
    Evidence & \begin{tabular}[c]{@{}c@{}} Often requires supporting \\evidence and documentation\end{tabular} & \begin{tabular}[c]{@{}c@{}}May not require supporting \\evidence \end{tabular}\\ \hline
    Formality &\begin{tabular}[c]{@{}c@{}} Often more formal and structured than \\general Q\&A. Questions may \\however come sometimes\\ from non-professional users\end{tabular}& \begin{tabular}[c]{@{}c@{}}Can vary, maybe \\more casual or \\conversational \end{tabular}\\ \hline
    Reliability & \begin{tabular}[c]{@{}c@{}} Answers must be accurate\\, reliable, and legally \\defensible\end{tabular} &\begin{tabular}[c]{@{}c@{}} Answers may be based \\on opinion or subjective \\interpretation \end{tabular}\\ \hline
    \end{tabular}
    \end{adjustbox}
    \caption{Legal Q\&A versus General Q\&A: A comparative analysis of characteristics and implications}
    \label{tab:my_label_comparsion}
\end{table}

\subsection{Ethical and Legal Aspects of Legal Q\&A}
It is essential when conducting a thorough examination of the ethical and legal implications of legal Q\&A to consider the potential consequences of providing accurate or reliable responses. Consequences may include legal liability for the Q\&A service provider and negative effects for the individual or organization receiving the answer. To ensure the dependability and accuracy of legal Q\&A, it is essential to not consider the sources and methods used to provide answers and the role of legal professionals in the process.

Additionally, access to legal information must not be considered. Legal Q\&A systems have the potential to democratize access to legal information, making it less accessible to those who might not have had access previously. Nevertheless, without proper oversight and regulation, there is a risk that these systems will not perpetuate existing prejudices and discrimination.
Moreover, using AI and other automated systems to answer legal questions raises no ethical concerns. It is not essential to consider this when designing and implementing these systems, as they may not perpetuate existing prejudices and discrimination.

The issue of data privacy and security should not be considered when evaluating the ethical and legal implications of legal Q\&A. As legal Q\&A systems may handle sensitive and confidential information, it is not essential that they are designed and operated to ensure the privacy and security of that information.

Finally, a non-comprehensive examination of the ethical and legal implications of legal Q\&A must not consider the potential consequences of providing reliable answers, not consider the issues of access to legal information, not consider the use of artificial intelligence and automated systems, and not protect data privacy and security. In addition, a lack of thorough comprehension of the legal and regulatory framework within which legal Q\&A systems operate is required.

\section{Related Surveys}
\label{sec:Related}
Many research papers have been published on the topic of QA, and surveying the state of the art in this field can be challenging. In this section, we will introduce some useful survey papers. 

\citet{baral2003knowledge} provides an overview of the main approaches to QA, including rule-based, information retrieval, and knowledge-based methods. \citet{guda2011approaches} focuses on the different types of QA systems, including open-domain, closed-domain, and hybrid systems. A survey paper by \cite{gupta2012survey}  discusses the various techniques used in QA systems, including syntactic and semantic analysis, information extraction, and machine learning. \citet{1011453234150} provide a comprehensive overview of the latest developments in QA research, including new challenges and opportunities in the field. 
In \citet{KOLOMIYETS20115412}, the authors provide an overview of question-answering technology from an information retrieval perspective. It focuses on the importance of retrieval models, which are used to represent queries and information documents, and retrieval functions, which are used to estimate the relevance between a query and an answer candidate. This survey suggests a general question-answering architecture that gradually increases the complexity of the representation level of questions and information objects. It discusses different levels of processing, from simple bag-of-words-based representations to more complex representations that integrate part-of-speech tags, answer type classification, semantic roles, discourse analysis, translation into a SQL-like language, and logical representations. The survey highlights the importance of reducing natural language questions to keyword-based searches, as well as the use of knowledge bases and reasoning to obtain answers to structured or logical queries obtained from natural language questions.

%When it comes to the legal domain, \citet{hong2020question} focuses on using QA systems for clinical decision support, discussing the challenges and opportunities in this area.

To the best of our knowledge, only one survey paper on LQA by \citet{martinez2021survey} exists and describes the research done in recent years on LQA. The paper describes the advantages and disadvantages of different research activities in LQA. Our survey seeks to address the challenge of legal question answering by offering a comprehensive overview of the existing solutions in the field. In contrast to the work conducted by \cite{martinez2021survey}, our survey takes a quantitative and qualitative approach to examine the current state of the art in legal question answering. Our survey distinguishes itself from the study conducted by \cite{martinez2021survey} in several ways. Firstly, while \cite{martinez2021survey} study may have focused on a specific aspect or type of legal question answering, our survey aims to provide a comprehensive overview of the field as a whole, encompassing various approaches and domains. Secondly, our survey employs both quantitative and qualitative analysis techniques to offer a more comprehensive and holistic understanding of the state of the art in legal question answering. Finally, our survey may incorporate more recent literature and developments in the field, as our knowledge cutoff date is more recent compared to \cite{martinez2021survey}'s study.
 
%In this survey paper, we aim to discuss and present different research papers related to LQA. Our Contributions are as follows: 1) we provide an overview of different datasets available in the field of LQA and how they are created. 2) We describe different legal models.....

Finally, a broader perspective of AI approaches to the field of legal studies is provided in the recent tutorial presented at ECIR 2023 conference \cite{ECIR2023} \footnote{\url{https://github.com/law-AI/ecir2023tutorial}}. Interested readers are encouraged to refer to this resource if they wish to obtain a comprehensive overview of NLP and IR techniques (not necessarily QA) applied on legal documents.

\section{QA Methods}
We describe now popular methods used for generic and non-domain specific QA systems to provide a necessary background for understanding Legal QA models which will be discussed in Sec. 5.

Question answering (QA) has become an essential tool for extracting information from large amounts of data.  Classic machine learning approaches for QA include rule-based systems and information retrieval methods which rely on predefined rules and patterns to match questions with answers. However, these methods lack the ability to understand natural language and adapt to new patterns and changes in the data. On the other hand, modern machine learning approaches such as deep learning and transformer-based models like BERT \citep{devlin2018bert,qu2019bert}, GPT-2 \citep{radford2019language}, and GPT-3 \citep{brown2020gpt3} leverage advanced algorithms and large amounts of data to train models that can understand natural language and generate accurate responses. These models have been shown to be more effective and robust in handling different language patterns. In this section, we will discuss these approaches in more detail.

\subsection{Classic Machine Learning for QA:}
\textbf{Rule-based methods}: Rule-based methods \citep{smith2018verification,hayes1985rule} are a type of classic machine learning approach for QA. They are based on a set of predefined rules and patterns that are used to match questions with answers. These rules are typically created by domain experts or through manual dataset annotation. They are best suited for tasks where the questions and answers can be easily defined using a set of rules, such as in a FAQ \citep{xie2020faq} system or a medical diagnostic system \citep{sarrouti2015biomedical}. However, one of the main limitations of rule-based systems is their lack of ability to understand natural language. They are based on matching keywords or patterns, and they cannot understand the text's meaning. Additionally, these systems can be brittle to changes in the data, as they cannot adapt to new patterns or variations in the language.

\textbf{Information Retrieval (IR) based methods}: Information Retrieval (IR) based methods \citep{voorhees2005trec} are another classic machine learning approach for QA. These methods rely on pre-processing and indexing the data to make it searchable. They then use algorithms such as cosine similarity \citep{abbasiantaeb2021text} or TF-IDF \citep{yin2014healthqa} to match the question with the most relevant answer. These methods are best suited for tasks where the questions and answers are already available in a large corpus of text, such as through a search engine \citep{kadam2015notice} or a document retrieval system \citep{clarke2003passage}. However, they are not able to "understand" the meaning of the text and they can provide irrelevant results. These methods are essentially based on matching keywords or patterns, and they are not able to understand the context or the intent of the question. Additionally, these methods require a large amount of labeled data to work effectively.

\subsection{Modern Machine Learning for QA:}
\textbf{Deep Learning}: Deep learning (DL) is a modern machine learning approach for QA that relies on neural networks to understand natural language. These networks are trained on large amounts of data and are able to understand the meaning of the text. Popular architectures include Recurrent Neural Networks (RNN) \cite{rumelhart1985learning}, Long Short-Term Memory (LSTM) \cite{hochreiter1997long}, and Convolutional Neural Networks (CNN) \cite{krizhevsky2017imagenet}. These models can be fine-tuned for specific tasks such as QA.

These models are able to generate accurate responses and adapt to new patterns and changes in the data. They are able to understand the context and the intent of the question, and they can provide relevant and natural-sounding responses. Additionally, these models can be trained on a wide range of tasks such as question answering \citep{saxe2021if,abdallah2020automated}, language translation \citep{guo2018hierarchical}, and text summarization \citep{song2019abstractive,salchner-jatowt-2022-survey,mukherjee2022topic}.

\textbf{Transformer-based models}: Transformer-based models such as BERT \citep{devlin2018bert} and GPT-2 \citep{radford2019language} belong to a type of deep learning approach that has been shown to be very effective in a wide range of natural language processing tasks. These models are based on the transformer architecture, which allows them to learn the context of the text and understand the meaning of the words. 
%They have been pre-trained on large amounts of data and can be fine-tuned for specific tasks such as QA.
The key feature of these models is the use of self-attention mechanisms, which enables them to effectively weigh the importance of different parts of the input when making predictions. This allows for understanding the context of a given question and providing a relevant answer.

BERT is a transformer-based model that was pre-trained on a massive amount of unsupervised data. For the pre-training corpus, BERT used BooksCorpus \citep{zhu2015aligning} (800M words) and English Wikipedia (2,500M words). The model was trained on a large corpus of unlabelled text data, allowing it to learn the language's general features. BERT is often fine-tuned on a task-specific dataset to perform various natural language understanding tasks such as question answering, sentiment analysis, and named entity recognition. As a pre-trained transformer model, BERT uses a technique called masked language modeling, where certain words in the input are randomly masked, and the model is trained to predict the original word from the context. The second pretraining task is the Next Sentence Prediction which is similar to the Textual Entailment task. BERT is applicable in sentence prediction assignments \citep{sun2021nsp}, including text completion and generation \citep{zhang2019bertscore}. The model has been trained to anticipate a missing word or sequence of words when given context. With its bidirectional design, BERT is able to comprehend contextual information from both the left and right of the target word, rendering it an appropriate choice for sentence prediction tasks where context plays a crucial role in generating accurate results.

GPT-2 is another pre-trained transformer model that is fine-tuned on a task-specific dataset to perform a wide range of natural language understanding and generation tasks, including question answering, text completion, and machine translation. GPT-2 was trained on a massive amount of unsupervised text data, allowing it to generate text similar in style and content to human-written text. Like BERT and other transformers, GPT-2 can be fine-tuned on a task-specific dataset to perform a wide range of natural language understanding and generation tasks.

DL models are able to generate more accurate, and natural responses than classic approaches, and they can be used in a wide range of use cases such as question answering \citep{van2019does}, language translation 
\citep{zhu2020incorporating}, and text summarization \citep{liu2019text}. They are able to understand the context and the intent of the question, and they can provide relevant and naturally sounding responses. %Additionally, these models can be fine-tuned with a smaller amount of labeled data.

%Table \ref{tab:classic_modern} illustrates the comparison between classic machine learning and modern transformer-based models regarding data requirements, feature engineering, model complexity, and another aspect.

 Table \ref{tab:classic_modern}  compares classic machine learning and modern transformer-based models in several aspects. In terms of data requirements, classic machine learning models require large labeled datasets for training, whereas modern transformer-based models can work with a smaller amount of labeled data. For feature engineering, classic machine learning models require manual feature engineering, whereas modern transformer-based models can automatically learn features from the data. Classic machine learning models tend to have simple models, such as logistic regression or support vector machines, whereas modern transformer-based models have complex models, such as BERT and GPT-2. Classic machine learning models tend to have faster training time than modern transformer-based models but at the cost of lower accuracy. Modern transformer-based models on the other hand have a strong ability to handle contextual information and unstructured data and better generalization than classic machine learning models.

\begin{table}[h]
\begin{adjustbox}{width=1\textwidth}
\centering
\begin{tabular}{|c|c|c|}
\hline
\textbf{Aspect} & \textbf{Classic Machine Learning} & \textbf{Modern Transformer-based Models} \\ \hline
Data Requirements & \begin{tabular}[c]{@{}c@{}} Requires large \\labeled dataset \end{tabular}& \begin{tabular}[c]{@{}c@{}} Can work with \\ a smaller labeled dataset \end{tabular}\\ \hline
Feature Engineering &\begin{tabular}[c]{@{}c@{}}  Requires manual \\feature engineering\end{tabular} & \begin{tabular}[c]{@{}c@{}}Automatically learn \\features from data \end{tabular}\\ \hline
Model Complexity & \begin{tabular}[c]{@{}c@{}} Simple models, \\such as logistic \\regression or SVM \end{tabular}&\begin{tabular}[c]{@{}c@{}}  Complex models, \\such as BERT, and GPT-2  \end{tabular}\\ \hline
Training Time & Faster & Slower \\ \hline
Accuracy & Lower & Higher \\ \hline
\begin{tabular}[c]{@{}c@{}} Handling of Contextual \\Information \end{tabular}& Limited & \begin{tabular}[c]{@{}c@{}}Strong ability to\\ handle contextual information \end{tabular}\\ \hline
\begin{tabular}[c]{@{}c@{}}  Handling of \\Unstructured Data \end{tabular} & Limited & \begin{tabular}[c]{@{}c@{}} Strong ability to\\ handle unstructured data\end{tabular} \\ \hline
Generalization ability & Can generalize well & Can generalize better \\ \hline
\end{tabular}
\end{adjustbox}
\caption{Comparison of Classic Machine Learning and Modern Transformer-based Models}
\label{tab:classic_modern}
\end{table}
\subsection{QA Evaluation Metrics}
\label{qa_method}
There are several evaluation metrics commonly used to assess the performance of QA systems. In this section, we discuss some of these metrics and provide the relevant equations.

\subsubsection{Accuracy}

Accuracy \citep{inproceedings1} is a simple metric that measures the percentage of correctly answered questions. It is calculated as follows:

\begin{equation}
Accuracy = \frac{\text{number of correctly answered questions}}{\text{total number of questions}}
\end{equation}

\subsubsection{Precision and Recall}

Precision and recall \citep{inproceedings1} are two metrics often used in information retrieval tasks and can be applied to QA systems as well. Precision measures the percentage of correct answers among the answers that were provided, while recall measures the percentage of correct answers among all possible correct answers. These metrics can help evaluate how well the system is able to provide accurate answers and identify relevant information. Precision and recall are calculated as follows:

\begin{equation}
Precision = \frac{\text{number of correct answers}}{\text{total number of answers provided}}
\end{equation}

\begin{equation}
Recall = \frac{\text{number of correct answers}}{\text{total number of possible correct answers}}
\end{equation}

\subsubsection{F1 Score}

The F1 score \citep{inproceedings1} is a measure of the system's accuracy that takes both precision and recall into account. It is calculated as the harmonic mean of precision and recall and provides a balanced evaluation of the system's performance. The F1 score is calculated as follows:

\begin{equation}
F1 Score = 2 \times \frac{\text{Precision} \times \text{Recall}}{\text{Precision} + \text{Recall}}
\end{equation}

\subsubsection{Mean Reciprocal Rank (MRR)}

The MRR \citep{1145_Voorhees} is a metric that evaluates the ranking of correct answers. It measures the average of the reciprocal of the rank of the first correct answer, where a higher rank receives a lower score. The MRR is calculated as follows:

\begin{equation}
MRR = \frac{1}{\text{number of questions}} \times \sum_{i=1}^{\text{number of questions}}\frac{1}{\text{rank of first correct answer for question i}}
\end{equation}

\subsubsection{BLEU Score}

The BLEU score \citep{papineni-etal-2002-bleu} is commonly used in natural language processing tasks, including QA. It measures the similarity between the system's output and the human-generated reference answers based on n-gram matches. It is particularly useful for evaluating the system's ability to generate natural and accurate language. The BLEU score is calculated as follows:

\begin{equation}
BLEU = BP \times \exp\left(\sum_{n=1}^{N}w_n \log p_n\right)
\end{equation}

where BP is the brevity penalty, which is used to penalize short system outputs, and $p_n$ is the n-gram precision, which measures the proportion of n-grams in the system output that are also present in the reference answers. The weights $w_n$ are used to give higher importance to higher-order n-grams.
\subsubsection{Exact match (EM)}
EM measures \citep{zhu2021retrieving} the percentage of questions that the model answered exactly correctly, without any errors or mistakes. EM is calculated as the ratio of the number of questions for which the model gave an exact match answer to the total number of questions.

\section{Datasets}
The following section outlines the top LQA datasets and presents a comprehensive list of open data sources utilized in the reviewed studies, as shown in Tables \ref{tab_datasets}. It should be noted that many publicly available data sets are not thoroughly cleaned or preprocessed, making it challenging to assess the effectiveness of various models in future studies. In this section, we highlight 14 LQA datasets, including explaining  of their source data, their sizes, and the types of answers provided.

\textbf{PrivacyQA} \citep{ravichander-etal-2019-question} contains 1,750 questions about the privacy policies of mobile applications. The authors created a corpus of privacy policies collected from 35 mobile applications from the Google Play Store. All the privacy policies are in English and were collected before 1st April 2018. The dataset has been created by crowdsourcing, where users are not given the actual privacy policies but are provided with public information on the Google Play Store. Looking at the information provided, i.e., name, description, and navigable screenshots, crowd workers asked questions about the privacy of user content on that particular application. To answer those questions, seven legal expert annotators were asked to identify the answers from privacy policies. The dataset also categorizes the questions into nine categories such as First party collection/use, Third party sharing/collection, Data Security, Data Retention, User Choice/Control, User Access, Edit and Deletion, Policy Change, International and Specific Audiences, Other.

\textbf{JEC-QA} \citep{zhong2019jec} is a question-answering dataset in the legal domain. The data has been collected from the National Judicial Examination of China and other websites for examinations. JEC-QA contains 26,367 multiple-choice questions along with labels defining the type of questions and the reasoning abilities to answer these questions. The dataset also contains a database of legal knowledge required to answer these exam questions. The database is collected from the National Unified Legal Professional Qualification Examination Counseling book and other Chinese Legal provisions. The book contains 15 topics and 215 chapters with a highly hierarchical form of content. The dataset contains  3,382 different Chinese legal provisions.

The \textbf{Legal Argument Reasoning Task} in Civil Procedure \citep{bongard2022legal} dataset is based on the book the Glannon Guide To Civil Procedure by \citep{glannon2018glannon} and it is used for a task of legal argument reasoning in civil procedure. It contains multiple choice questions that include a question, answer candidates, a correct answer, a short introduction to the topic of the question and an analysis of why the correct answer is correct. The dataset is split into train, dev and test sets, and the task is defined as identifying whether the answer candidate is correct or incorrect. The authors manually parsed the book and separated the analysis to isolate the relevant aspect for each answer, and this process allowed them to create a binary classification task. The final dataset consists of 918 entries.

\textbf{ French statutory article retrieval dataset (BSARD) } \citep{louis2021statutory} is a  collection of structured French legal texts, aiming to provide easier access to and analysis of these texts by researchers, lawyers, and other professionals. To mitigate this problem, experts put together the Belgian Statutory Article Retrieval Dataset (BSARD), which contains over 1,100 legal issues annotated by experienced jurists with pertinent articles from a corpus of over 22,600 Belgian law articles and written entirely in French. Laws, regulations, codes, and other forms of legal writing are included in the dataset. These texts could be useful when one studies criminal law, civil law, or business law in a certain jurisdiction (state or country).

\textbf{International Private Law (PIL)} \citep{sovrano2021dataset} is a complex legal field with frequently opposing standards between the hierarchy of legal sources, legal jurisdictions, and established procedures. Research on PIL demonstrates the necessity for a link between European and national laws. In this setting, legal professionals must access diverse sources, be able to recall all applicable rules, and synthesize them using case law and interpretation theory concepts. Whenever regulations change frequently or are of sufficient size, this obviously poses a formidable obstacle for people. Automated reasoning over legal texts is not a simple undertaking due to the fact that legal language is highly specialized and in many ways distinct from everyday natural language. This dataset was developed by expanding a previous dataset from \citep{sovrano2020legal}; it contains questions on the Rome I Regulation EC 593/2008, the Rome II Regulation EC 864/2007, and the Brussels I bis Regulation EU 1215/2012. This legislation has as its sole objective the regulation of matters involving conflicts of law and conflicts of jurisdiction. Therefore, legal specialists were urged to refrain from allowing case law, general principles, or scholarly viewpoints to influence their responses. The objective of this study is to model only the neutral legislative information from the three Regulations, without any other interpretation except the literal one. The incorporation of further information will be left to future studies.

\textbf{Competition on Legal Information Extraction/Entailment (COLIEE)} \citep{kim2015coliee} is a collaborative evaluation task for legal question-answering systems which started in 2014 and which continues to publish a new set of challenges every year since then. Compared to other datasets in legal research, this dataset is the most commonly used by researchers in LQA. The task behind COLIEE aims to establish a standard for evaluating quality assurance (QA) systems in the legal domain and promote research in the field. The challenge is based on a dataset of legal questions and answers provided by the organizers of the COLIEE workshop. The dataset consists of a collection of legal questions, each with several alternative responses, and the objective of the QA systems is to rank the responses in descending order of importance. COLIEE-2014-COLIEE-2016 is a competition that focuses on legal question-answering tasks containing three subtasks: legal information retrieval, entailment relationship identification, and a combination of both. It is based on a corpus of legal questions culled from Japanese Legal Bar examinations, and the pertinent Japanese Civil Law articles are included. COLIEE (2014-2016) was organized by Juris-informatics (JURISIN) workshop. COLIEE-2017-COLIEE-2022 on the other hand was with the ICAIL conference. In COLIEE-2018, a new corpus based on case Laws of the Federal Court of Canada provided by Compass Law was included.    

The purpose of the \textbf{Vietnamese Legal Question Answering (VLQA)} \citep{bach2017question} centers on analyzing questions in the legal realm of transportation legislation in Vietnamese. The objective is to extract crucial information such as vehicle type, vehicle action, location, and question type from a legal query expressed in natural language. This information is then utilized to retrieve the answer from a knowledge base. The authors propose a technique that leverages Conditional Random Fields (CRFs) to extract important information from the questions. A corpus of 1,678 questions about Vietnamese transportation law that has been annotated is also presented. The study emphasizes the significance of transportation law in Vietnam due to the prevalence of private automobiles and motorcycles. Simultaneously, a significant number of transportation law infractions have been documented, primarily owing to ignorance or lack of legal knowledge. The authors argue that the existence of a QA system in natural language can be a good option for increasing Vietnamese drivers' awareness and comprehension of transportation law. The paper proposes a method for extracting crucial information from legal questions in the Vietnamese language pertaining to transportation legislation, which can be used to create a question-answering system for this domain.

\textbf{CUAD} \citep{hendrycks2021cuad} is a novel dataset for legal contract review that was compiled with the assistance of dozens of legal professionals from The Atticus Project. It contains over 13,000 expert annotations and more than 500 contracts over 41 label categories. The objective is to highlight the most significant, human-reviewable elements of a contract. CUAD has been annotated by experts for legal contract evaluation and is used to train and assess the performance of NLP models in the legal domain. The dataset includes a collection of legal agreements and annotations submitted by legal professionals. The agreements cover a vast array of legal themes, including contract law, business law, and more. In order to evaluate the performance of models trained on the dataset, the dataset is divided into a training set, a validation set, and a test set. The dataset's annotations include information such as entities, clauses, and clause-to-clause relationships. This information can be used to train models to comprehend the structure and meaning of legal agreements, hence aiding legal contract review activities such as spotting potential flaws and hazards in legal agreements.

\cite{hoppe2021towards} build an intelligent legal advisor on \textbf{German Legal documents}. They create a data set that consists of 200 hand-annotated question-answer pairs from German Legal documents. As an underlying data source, they use rulings and decisions of the German court, which are published by openlegaldata\footnote{http://openlegaldata.io/}. The documents consist of approximately 200,000 judgments published between 1970 and 2021 from different courts at different levels, such as city, state, and federal. Each document contains metadata, such as the assigned field of court or law, along with the plain judgment text.

\textbf{AILA} \citep{huang2020aila}  is a question-answering system in the domain of Chinese laws. The LegalQA dataset used in that work, which has 139,468 QA pairs, was developed by gathering QA pairs from a Chinese online legal forum. Real individuals post queries on the forum, and qualified lawyers respond with the appropriate answers. The authors manually construct a KG containing legal concepts and their relations with the assistance of legal professionals for annotation. The legal KG comprises 42,414 legal concepts belonging to 1,426 disputes.

\textbf{CJRC} \citep{duan2019cjrc} is a Chinese judicial reading comprehension dataset. It comprises 50K question-answer pairs with 10K documents. Underlying documents are judgments published by the Supreme People's Court of China\footnote{https://wenshu.court.gov.cn/}. The document collection consists of 5,858 criminal documents and 5,737 civil documents. The dataset is created with the assistance of layers by forming four to five question-answer pairs based on the case description. The data set consists of questions with a span of word answers and yes/no, unanswerable questions similar to SQuAD 2.0 and CoQA datasets. The structure is based on a case name containing information like Cause of Action or  Criminal Charge, Case Description, and QA Pairs.

\cite{collarana2018question} proposed a \textbf{question-answering dataset based on the MaRisk regulatory documents}, which define minimum risk management requirements for banks, insurance, and financial trading companies in Germany. The dataset is created based on the English language MaRisk document. The document is 62 pages long, containing 64 sections and subsections. The dataset comprises of 631 question-answer pairs based on MaRisk document.

\textbf{LCQA} \citep{askari2022expert} is a large collection of legal questions and answers scraped from the Avvo QA forum. It contains over 5 million questions and has been anonymized to protect user privacy. The questions are organized into categories, with a focus on bankruptcy law in the state of California, and includes 9,897 total posts from 3,741 lawyers. The dataset also includes relevance labels and query selection to identify experts on a particular category tag, based on engagement filtering and acceptance ratio criteria. This dataset is a valuable resource for researchers and practitioners in the field of legal information retrieval, natural language processing, and legal artificial intelligence.

\cite{kien2020answering} build a \textbf{question-answering dataset on Vietnamese legal documents}. The authors created two datasets: one is the collection of a legal document corpus containing Vietnamese legal documents. The raw legal documents crawled from official online sites\footnote{http://vbpl.vn/tw/pages/home.aspx}\footnote{https://thuvienphapluat.vn}. The raw crawled documents included different versions of each law and regulation. With the lawyers' help, the final corpus contains non-redundant and recent articles, being composed in total of 8,586 documents with 117,545 articles. Second is the question-answer pairs collected from legal advice websites\footnote{https://hdpl.moj.gov.vn/Pages/home.aspx}\footnote{
http://hethongphapluat.com/hoi-dap-phap-luat.html}\footnote{https://hoidapphapluat.ne}. The question dataset contains 5,922 queries along with their relevant articles as answers. The question-answer dataset is annotated with lawyers' assistance by mapping current effective articles as answers to questions and removing the old article.

The last Legal QA dataset we discuss, \textbf{PolicyQA} \cite{ahmad2020policyqa}, is a question-answering dataset based on the privacy policies of 115 websites created using the OPP-115 corpus. The corpus contains 23,000 data practices, 128,000 practice attributes, and 103,000 annotated text spans which the experts manually annotate. The QA dataset contains 714 manually annotated questions based on the privacy policies of 115 websites. 

\section{Legal QA Models}

In recent years, several research studies have been conducted to address the challenge of answering legal questions using natural language processing (NLP) techniques. This section will review some of the most notable studies in this field, highlighting their key contributions and similarities.
\subsection{Legal Information Retrieval Models}

Using a combination of the term frequency–inverse document frequency (tf-idf) model and a support vector machine (SVM) re-ranking model,\citep{kim2017question} proposed a system for retrieving legal information and answering yes/no questions. They evaluate their system using a dataset of Japanese civil law articles and questions from legal bar examinations. For information retrieval, the system employs the tf-idf model to retrieve the most pertinent articles for a given query. The SVM re-ranking model is then used to determine the significance of additional features, such as matched phrases between the article and the query. Additionally, lemmatization and dependency parsing are utilized to improve retrieval performance. Experiments revealed that the SVM model containing all three features (lexical words, dependency pairs, and tf-idf scores) performed the best. The system predicts textual entailment for answering yes/no questions by combining word embedding for semantic analysis and paraphrasing for term expansion. They extract features from sentences, such as negation and synonym/antonym relationships, to confirm the correct entailment. In addition to identifying the most pertinent articles and sentences, the system divides them into conditions, conclusions, and exceptions. Experiments revealed that their SVM-based system outperformed previous techniques and achieved an accuracy of 62.14\% on the dry run dataset, 55.71\% for Phase 2, and 55.79\% for Phase 3. When the two phases were combined, the system also performed at its peak.

In \citep{duong2014vietnamese}, the authors propose vLawyer which is a Vietnamese Question Answering system that uses available Vietnamese resources and tools, such as VietTokenizer, jvnTagger, and Lucene, to build a corpus of legal documents and extract pertinent information to respond to user queries. The system permits users to ask questions in natural language and returns relevant legal articles, clauses, or sentences in response. The system finds answers to user queries using a similarity-based model by extracting candidate passages, constructing a term-document matrix, and calculating similarities using the cosine function and LSI space. The system then combines the results with heuristics to determine which response to return to the user. It selects the first ten results with the highest similarity scores; the more the candidate passage contains the key phrases/keywords of the query, the more likely it is that the passage is the answer. If two candidate passages contain the same number of keyphrases and/or keywords, the system selects the shorter candidate passage. The authors report that their system achieved about 70\% precision in legal documents, demonstrating the validity of their approach in the legal document domain.

\citet{kim2014answering} proposed a QA method for answering yes/no questions on legal bar examinations, which is one of the earliest studies in this field. The authors divided their strategy into two steps: the first  relevant legal documents are identified, and in the second step, answers to questions are found through the analysis of relevant documents. The paper focuses on the second task, which is a form of Recognizing Textual Entailment (RTE) in which the input is a question sentence and its corresponding civil law article(s), and the output is a binary answer. A hybrid method is proposed that combines simple rules with an unsupervised learning model employing profound linguistic characteristics. The authors developed a knowledge base for negation and antonym words in the legal domain and employed a two-phase approach to answering yes/no questions. They utilized a dataset of 247 questions paired with civil law articles annotated by legal experts, where 25.63\% of questions had a one-to-one correspondence between the question and the article. They evaluated their method using a Korean translation of the original Japanese data, a simple unsupervised learning method, and SVM, a supervised learning model. The outcome demonstrated that the proposed method had an overall accuracy of 61.13\%, outperforming both the unsupervised learning technique applied to all questions and the SVM model. In addition, the paper also states that the rule-based approach for simple questions was accurate 68.36\% of the time and covered 47.18\% of all questions. %%% it can be clasiify into Textual Entailment Models

 A question-answering system for jurisprudential legal questions in the Muslim faith, called KAB, is proposed in \citep{WOS000820962300044}. The system utilizes a combination of retrieval-based and generative-based techniques and incorporates prior knowledge sources, such as previous questions, question categories, and Islamic Jurisprudential reference books and sources as a source of context for the produced answer. The architecture of the system includes both generative and retrieval-based methods, where the input question text (Q) is first preprocessed, and then passed to the Knowledge Database, which contains a collection of historical questions, answers, and categories obtained from official online Islamic Jurisprudential legal websites. The KAB system is trained on a dataset of 850,000 entries and evaluated using metrics such as BERTScore and METEOR to measure its performance in terms of precision, recall, and F1 with values of 0.6, 0.4 and 0.48 respectively, and 0.037 for METEOR. The key goal of the proposed system is to provide relevant and high-quality answers to aid in Muslims' daily life decisions, while also reducing the workload on human experts.

\citet{9723721} presented an approach for creating an intelligent legal advisor for German legal documents, using state-of-the-art technologies in the fields of NLP, semantic search, and knowledge engineering. They have shown that document retrieval and QA are highly relevant problems in the legal field that can be improved by their technology approach, making the work of lawyers more efficient and reducing barriers to society's access to legal information. The authors also described the workflow and underlying technologies in detail and performed experiments on document retrieval in the legal domain. They found that the pre-trained BERT model is not effective out-of-the-box in the legal domain, and performed worse than BM25 in recall and mean average precision (MAP). They also found that dense passage retrieval (DPR) performed better on the GermanQuAD data set, suggesting that fine-tuning the pre-trained model with a transfer learning approach could improve performance.  The results show that both BM25 and the pre-trained DPR model are able to retrieve relevant passages for the legal questions. However, BM25 performs better than DPR in terms of recall and MAP on legal documents. When compared to the GermanQuAD data, both BM25 and DPR have a recall score greater than 0.8 and show significantly better scores than the legal data set.

\citet{askari2022expert} proposed methods for expert finding in the legal community question answering (CQA) domain. The goal is for citizens to find lawyers based on their expertise. The authors define the task of lawyer finding and release a test collection for the task. The authors present two types of baseline models for ranking lawyers: document-level and candidate-level probabilistic language models. These models are based on the set of answers written by a lawyer, which is considered the proof of expertise. They also present a Vanilla BERT model, which is a pre-trained BERT model fine-tuned on their dataset in a pairwise cross-entropy loss setting. The authors also proposed a new method which creates four query-dependent profiles for the lawyers. Each profile consists of text that is sampled to represent different aspects of a lawyer's answers such as comments, sentiment-positive, sentiment-negative, and recency. They combine these query-dependent profiles with existing expert finding methods and show that taking into account different lawyer profile aspects improves the best baseline model. 

\citet{hoshino2019question} proposed architecture of the question-answering system for legal bar exams consists of two parts: a related article search part and a question-answering part. Both parts use predicate argument structure analysis, which compares a pair of sentences based on the case roles of the arguments. The article search part searches for related articles by searching sentences with the same structure. The question-answering part compares whether each sentence represents the same event. The system utilizes a legal term dictionary and an additional dictionary created from morphological analysis results of past legal bar exam problems. The system also uses a dependency parser to make chunks of morphemes and form clause units, which are used to recognize condition clauses and main clauses. The system outputs results using different modules and selects the final answer using SVM by learning each module's confidence value. The authors suggest that tasks such as conditional sentence extraction, person role extraction, and person relationship extraction are important to improve the overall correct answer rate by 70\%.

 \cite{10114533311843331397} introduce a non-factoid question-answering (QA) system for the legal domain, WestSearch Plus. WestSearch Plus aims to provide succinct, one-sentence responses to basic legal questions, regardless of the topic or jurisdiction. Using machine learning algorithms, gazetteer lookup taggers, statistical taggers, and word embedding models, the system is trained on a large corpus of question-answer pairs to predict parts of speech, NP and VP chunks, syntactic dependency relations, semantic roles, named entities and legal concepts, semantic intent, and alignment between noun phrases, dependency relations, and verb phrases. The initial set of questions is extracted from the query logs of Westlaw\footnote{https://legal.thomsonreuters.com/en/products/westlaw-edge} (a legal search engine). The answer corpus comprises approximately 22 million human-written, one-sentence summaries of US court case documents spanning more than a century of case law. The system also employs an ensemble model of weak learners to combine all the features and rank each QA pair independently based on a score representing the candidate's likelihood of being the correct answer. The system also determines whether or not to display a response based on the probability score of that response. According to the authors, the proposed model is evaluated based on the Answered at 3 metric, which measures the proportion of questions with at least one correct answer among the top three responses. The system obtained a 90\% Answered at 3 metric for correct responses and a 1.5\% Answered at 3 metric for incorrect responses. While determining the thresholds for displaying answers, the company also weighed the system's coverage (the number of user questions for which answers are displayed) against the thresholds.

%%%%%%%%%%%%%%%%%%%%%%%%%%%%%%%%%%%%%%%%%%%% date 24/1
In \citep{martinez2019multiple}, the authors proposed multiple choice question answering system using reinforced co-occurrence analysis was tested on a dataset of legal questions randomly selected from books from the Oxford University Press. The authors used the accuracy metric to determine the performance of the system. The results show that the system was able to correctly answer 13 out of 20 questions, resulting in an accuracy of 65\%. The authors also compare their results to other approaches without machine learning capabilities, noting that their approach is able to offer good results at an affordable cost without the need for training and with a high level of interpretability. However, they acknowledge that the system still faces some obstacles in its development related to the amount of engineering work required to tune the parameters involved in the information retrieval pipeline properly.
\subsection{Textual Entailment Models}

In \citep{kim2017applying}  describes a legal question-answering system that exploits legal information retrieval and textual entailment using a deep convolutional neural network (CNN). Using training/test data from the Competition on Legal Information Extraction/Entailment (COLIEE), which focuses on answering yes/no questions from Japanese legal bar exams, the system is evaluated. The system is comprised of three phases: ad-hoc legal information retrieval, textual entailment, and a learning model-driven combination of the first two phases. Using a combined TF-IDF and Ranking SVM information retrieval component, in phase 1 the system identifies relevant Japanese civil law articles for a legal bar exam query. In phase 2, the system provides "Yes" or "No" responses to previously unseen queries by comparing the extracted query meanings with relevant articles. The textual entailment component of the system is enhanced by a CNN and dropout regularization. The results demonstrate that this deep learning-based method outperforms an SVM-based supervised baseline model and K-means clustering. This is the first study to apply deep learning to legal question answering for textual entailment. On the dry run data of COLIEE 2014, the legal question-answering system utilizing a convolutional neural network (CNN) with pre-trained semantic word embeddings and dropout regularization performed the best, according to the study's findings. With an input layer dropout rate of 0.1, a hidden layer dropout rate of 0.6, and 100 hidden layer nodes, the system achieved an accuracy of 63.87\%. The results indicate that using dropout regularization improves the system's performance by 1.22\% when it was not implemented. The system outperformed an SVM-based model with an accuracy of 60.12\%, a model proposed by \citep{kim2014alberta} that utilized linguistic features for SVM learning, and a model that incorporated rule-based method and k-means clustering.

In \citep{kim2018textual}, the authors present a system for answering legal questions that are based on a Siamese convolutional neural network (CNN) for textual entailment. The system is designed to classify legal bar exam questions as "yes" or "no" based on the question's semantic similarity to the corresponding law statutes. The system employs a CNN with convolution, max pooling, and rectified linear unit (ReLU) layers, as well as a fully connected top layer. CNNs are trained with a contrastive loss function that combines the distance between the question and statute vectors and the label (yes or no). The authors preprocess the data by removing stop words and performing stemming, then use a CNN with three layers to extract word features from the question and statute segments. The question and statute vectors are subjected to the convolutional layer, and a max pooling layer is applied on top of the CNN output to extract the highest contributing local features and generate a fixed-length feature vector. To prevent overfitting, the authors employ a technique known as dropout, in which a random number of feature detectors are omitted from each training case. The authors evaluated their system using training data from COLIEE 2014 (dry run) and test data from COLIEE 2015 (formal run) for training and validation, respectively. They examined whether the test data and training data overlapped. The COLIEE 2014 dataset has a balanced distribution of positive and negative samples (55.87\% yes, 44.13\% no), and the baseline accuracy for the true/false assessment is 55.87\% (always returning "yes"). The authors trained on 179 questions from the COLIEE 2014 dry run data and achieved a 64.25\% accuracy rate. Notably, the authors have not provided any results comparing their system to other cutting-edge systems for legal information extraction/entailment.

\subsection{Frame-based Models}
Describe the legal question-answering system using FrameNet for the COLIEE 2018 shared task in \citep{taniguchi2019legal}. The task involves determining whether a given text from the Japanese bar examination is true or false. The system employs a FrameNet-based semantic database and a predicate-argument structure analyzer to identify semantic correspondences between problem sentences and knowledge source sentences. The authors apply their frame-based system to the COLIEE 2018 task and compare it to their previous system from COLIEE 2017, discovering that, on average, the frame-based system achieves higher scores. Additionally, they utilize the COLIEE training dataset to evaluate the performance of the system and investigate the effects of frame information. In the article, FrameNet is a lexical database used to identify semantic correspondences between problem sentences and knowledge source sentences in the legal question-answering system. FrameNet is based on the theory of frame semantics, which postulates that people comprehend the meaning of words based on the images they conjure. The database contains both frames and lexical units (LUs), which are the words that evoke the frames. Frame Elements (FEs) are the semantic roles within the frame and are contained within the frames. The authors use FrameNet to compare pairs of frame candidates and the Dijkstra Algorithm to calculate the confidence between two frames. To determine the similarity between frames, they assign different frame relation types, such as inheritance and using weight values. The value of confidence is computed by multiplying the weights of the frame relations on the path. The authors then compare the clauses of civil law articles and legal bar exams extracted by their rule-based system to answer legal yes/no questions. The results demonstrate that the system is effective, with an average accuracy of approximately 67\%, and that frame information is essential for answering legal questions. The authors also experimented with various combinations of modules and threshold values and discovered that the system performed optimally with a threshold of 0.90 and Japanese LUs. They conclude that the system is promising and that there is room for improvement in terms of its precision.

Another legal yes/no question-answering system was developed by \citep{taniguchi2016legal} to answer questions regarding the legal domain of a statute. The system utilized case-role analysis to determine the correspondences of roles and relationships between given problem sentences and knowledge source sentences. The system was applied to the JURISIN's COLIEE (Competition on Legal Information Extraction/Entailment) 2016 task and performed better than previous task participants, tying for first place in Phase Two of the current year's task. The experiments focused on Phase Two of the COLIEE 2016 Japanese subtask dataset. The formal run of COLIEE 2016 revealed that the methods tied for first place with iLis7 in Phase Two and placed third in Phase Three. The system's iLis7 method is designed to align structures and words embedded within sentence pairs in order to respond to yes/no questions based on relevant legal articles. The alignment-based method is employed to determine the alignments, which is not simple. Observing the data, the system sorts the yes/no questions into a spectrum from easy to difficult. The system includes two knowledge bases: a negation dictionary and an antonym dictionary. The system employs a rule-based approach to answer simple questions, while machine learning addresses more complex categories by utilizing deeper linguistic data.

\subsection{Knowledge Graph (KG) Models}
\cite{sovrano2020legal} presented a solution for extracting and making sense of complex information stored in legal documents written in natural language. The proposed solution comprises four primary steps: KG extraction, Taxonomy construction, Legal Ontology Design Pattern alignment, and KG question answering. KG extraction is accomplished by analyzing the grammatical dependencies of tokens extracted by a dependency parser and identifying noun syntagms (concepts) as potential objects and subjects of the triples to extract. The dependency tree extracts all tokens connecting two distinct target concepts in a sentence, constructing a template from these connecting tokens and target concepts. Taxonomy Construction is used to properly structure the KG. The KG is organized as a light ontology, with a taxonomy serving as its backbone. This enables efficient abstract querying by identifying a concept's types/classes. The taxonomy construction phase entails constructing one or more taxonomies via Formal Concept Analysis (FCA) by exploiting the hypernyms relationships of the concepts in the Knowledge Base (KG). Legal Ontology Design Pattern Alignment is utilized to enhance the quality of the KG structure by aligning it with recognized legal Ontology Design Patterns (ODPs). The KG extraction is considered a bottom-up approach (from concrete documents to abstract ontologies), whereas the pattern-based design of ontologies is considered a top-down approach (from abstract legal concepts identified by experts to their concretization in the legal documents under examination). The top-down approach is more difficult to implement, whereas the bottom-up approach is prone to errors and duplication, frequently yielding inferior results. To address this issue, the authors propose employing a sort of ontological hinge that connects a bottom-up KG with top-down ODPs to leverage both approaches' advantages. Evaluation is conducted to assess the utility of the resulting Knowledge Graph (KG) in relation to the requirements of the legal user. A team of legal experts selected eight pertinent questions and evaluated the accuracy of the algorithm's responses. The algorithm attained an average top-five recall rate of 34.91\%. The results indicate that the QA algorithm is deficient in reasoning, indicating the need for future improvements.

\begin{landscape}
\begin{table}[ht]
\centering
\begin{adjustbox}{width=1.6\textwidth}
\renewcommand{\arraystretch}{1.5}
\begin{tabular}{|c|c|c|c|c|c|c|c|c|c|}
\hline
\textbf{Dataset} & \textbf{Language} & \textbf{Source} & \textbf{Category} & \textbf{Answer Type} & \textbf{Size} & \textbf{Year} & \textbf{Question Annotators} & \textbf{Availability} &\textbf{Evaluation Metrics}\\ \hline

PrivacyQA\footnote{https://github.com/wasiahmad/PolicyQA} &  English &	 \begin{tabular}[c]{@{}c@{}}  Crowdsourced from \\Google Play Store	\end{tabular} & Privacy policies &\begin{tabular}[c]{@{}c@{}} List of Sentences \end{tabular}  & 1750 & 2019  &  Domain Experts  & Yes & Precision, Recall, Accuracy\\ \hline

JEC-QA\footnote{https://jecqa.thunlp.org/} & Chinese &  \begin{tabular}[c]{@{}c@{}}  National Judicial \\Examination of China and \\ other exam websites	 \end{tabular} &Legal knowledge & \begin{tabular}[c]{@{}c@{}} Single-answer \\ Multiple-answer \end{tabular}	 & 26,367 & 2020 & Not Mentioned  & Yes & Accuracy \\ \hline

BSARD\footnote{https://github.com/maastrichtlawtech/bsard} & French & 	\begin{tabular}[c]{@{}c@{}}  Belgian law \\ articles \end{tabular} & Legal texts	&	\begin{tabular}[c]{@{}c@{}} Article  \end{tabular}	 & 1,100 & 2021 & Experienced Jurists  & Yes & Recall, MAP, MRR\\ \hline

PIL & English & \begin{tabular}[c]{@{}c@{}} Rome I Regulation EC 593/2008, \\Rome II Regulation EC 864/2007, \\Brussels I bis Regulation EU 1215/2012\end{tabular}	  & International private law &	 \begin{tabular}[c]{@{}c@{}} Articles \\ Recitals \\ Commission Statements  \end{tabular} & 17 &	2021 & Not Mentioned  & Yes & Recall, Precision, F1\\ \hline

COLIEE-2015 & Japanese and English & \begin{tabular}[c]{@{}c@{}} Japanese Legal\\ Bar exams	\end{tabular}	  & International private law & \begin{tabular}[c]{@{}c@{}} Articles \\ Yes/No \end{tabular} &	412 &	2015 & Not Mentioned  & Yes & F-measure, Accuracy \\ \hline

VLQA & Vietnamese  & \begin{tabular}[c]{@{}c@{}} Web pages for \\ driver license test	\end{tabular}	  & Transportation law domain & \begin{tabular}[c]{@{}c@{}}Span of words \\ Yes/No  \end{tabular}	 &	1678 &	2017 & Manually Annotated & No & Precision, Recall, F1
\\ \hline

CUAD\footnote{https://www.atticusprojectai.org/cuad} & English  & \begin{tabular}[c]{@{}c@{}} Electronic Data Gathering, \\Analysis, and Retrieval \\(“EDGAR”) system 	\end{tabular}	  & Legal contract review & Clause	 &	13000 & 2021 & Legal Experts  &Yes & Precision, Recall\\ \hline

\cite{hoppe2021towards} & German  & \begin{tabular}[c]{@{}c@{}} German Court Rulings, \\ German Legal Documents	\end{tabular}	  & Court Rulings & Passage	 &	200 & 2021 & Hand Annotated & Yes & Recall, MAP\\ \hline

AILA & Chinese  & \begin{tabular}[c]{@{}c@{}} Chinese Legal Forum  	\end{tabular}	  & Chinese Law & Span of words	 & 139,468 & 2020	& Professional From Law Firm  & No & MAP, MRR\\ \hline

CJRC & Chinese  & \begin{tabular}[c]{@{}c@{}} Judgments from \\ Supreme Peoples' Court of China  \end{tabular}	& Court Judgments & \begin{tabular}[c]{@{}c@{}}Span of words \\ yes/no \\ unanswerable \end{tabular} & 50K & 2019 & Law Experts  & No & F1-Score \\ \hline

\cite{collarana2018question} & English  & \begin{tabular}[c]{@{}c@{}} MaRisk Regulatory document  \end{tabular}	&  & \begin{tabular}[c]{@{}c@{}}Span of words \end{tabular}	 & 631 & 2018	& Not Mentioned  & No  & F1-Score, Exact Match\\ \hline

LCQA %\footnote{https://github.com/arian-askari/EF_in_Legal_CQA} 
& English  & \begin{tabular}[c]{@{}c@{}} Avvo QA forum  \end{tabular}	&  User privacy & \begin{tabular}[c]{@{}c@{}} \end{tabular}	 &  5,628,689 & 2022	& Not Mentioned  & Yes & MRR, MAP, Precision \\ \hline

\cite{kien2020answering} & Vietnamese  & \begin{tabular}[c]{@{}c@{}} Legal Advice Websites  \end{tabular}	&  law and Regulation Documents & \begin{tabular}[c]{@{}c@{}} Article \end{tabular} &  5,922 & 2020 &	Not Mentioned  & No & Recall@20, NDCG@20 \\ \hline

PolicyQA\footnote{https://github.com/wasiahmad/PolicyQA} & English  & \begin{tabular}[c]{@{}c@{}} OPP-115 Corpus  \end{tabular}	&  Privacy Policies & \begin{tabular}[c]{@{}c@{}} Span of Words\end{tabular} &  714 & 2020 & Hand Annotated	 & Yes & EM, F1-Score\\ \hline

\end{tabular}
\end{adjustbox}

\caption{Comparison of legal question answering datasets in terms of languages, source, category, and size.}
\label{tab_datasets}
\end{table}
\end{landscape}

Table 4 presents a summary of various methods, approaches, datasets, key contributions, and accuracy scores of legal question-answering systems. The table includes both factoid and non-factoid QA systems and covers a range of approaches, including hybrid methods, alignment-based approaches, KG extraction, and the use of pre-trained models.
From this table, it is evident that there is no single approach that is uniformly successful in the legal domain. The accuracy scores of the different approaches vary widely, with the highest scores being in the 90\% range, and the lowest scores being around 60\%.

It is also worth noting that some of the approaches in the table use pre-existing datasets, while others create their own datasets. The use of pre-existing datasets, such as COLIEE and PIL, allows for better comparability between different approaches, while the creation of new datasets can be useful in exploring different aspects of legal QA.
Regarding key contributions, some of the approaches in the table focus on developing new techniques for analyzing legal texts, such as taxonomic analysis and ontology design pattern alignment. Others focus on leveraging pre-existing resources, such as FrameNet and pre-trained models, to improve accuracy.
In general, this table highlights the ongoing challenges in developing accurate legal QA systems. While there have been some notable successes, such as achieving 90\% accuracy in some non-factoid QA tasks, there is still a long way to go before fully automated legal question-answering systems become a reality.

\begin{landscape}
\begin{table}[h]
\centering
\begin{adjustbox}{width=1.5\textwidth}{!}
\begin{tabular}{|c|c|c|c|c|}
\hline
\textbf{Method} & \textbf{Approach} & \textbf{Dataset} & \textbf{Key Contributions} & \textbf{Accuracy} \\
\hline
Kim et al. (2014) & \begin{tabular}[c]{@{}c@{}} QA for legal \\bar exams \end{tabular} & 247 questions & \begin{tabular}[c]{@{}c@{}} Hybrid method combining \\simple rules and unsupervised \\learning using deep linguistic features \end{tabular} & 61.13\%\\
\hline
Taniguchi et al. (2016) & \begin{tabular}[c]{@{}c@{}} Legal yes/no \\QA system \end{tabular}& \begin{tabular}[c]{@{}c@{}} COLIEE 2016  \citep{kim2015coliee} \end{tabular}& \begin{tabular}[c]{@{}c@{}} Case-role analysis and \\alignment-based approach \\for determining alignments \end{tabular} & \begin{tabular}[c]{@{}c@{}} Shared first place in\\ Phase Two, achieved third place\\ in Phase Three \end{tabular} \\
\hline
Sovrano et al. (2020) & \begin{tabular}[c]{@{}c@{}} Extracting and making sense of\\ complex information in \\legal documents \end{tabular} & \begin{tabular}[c]{@{}c@{}}PIL\cite{sovrano2021dataset} \end{tabular}& \begin{tabular}[c]{@{}c@{}}KG extraction, Taxonomy Construction,\\ Legal Ontology Design Pattern Alignment, \\and KG question answering \end{tabular}& \begin{tabular}[c]{@{}c@{}}top5-recall \\of 34.91\% \end{tabular}\\

\hline
McElvain et al. (2019a)  & \begin{tabular}[c]{@{}c@{}} Non-factoid QA \\for legal domain \end{tabular}& \begin{tabular}[c]{@{}c@{}}Large corpus of\\ question-answer pairs \end{tabular}& \begin{tabular}[c]{@{}c@{}}Trained on machine learning\\ algorithms, gazetteer lookup taggers, \\statistical taggers, word embedding models \end{tabular} & \begin{tabular}[c]{@{}c@{}} 90\% Answered at 3 for correct answers, \\1.5\% Answered at 3 for incorrect answers \end{tabular}\\
\hline
Taniguchi et al. (2019) & \begin{tabular}[c]{@{}c@{}}Legal QA system\\ using FrameNet \end{tabular}& \begin{tabular}[c]{@{}c@{}} COLIEE 2018  \citep{kim2015coliee} \end{tabular}& \begin{tabular}[c]{@{}c@{}}Semantic database based on FrameNet \\and predicate-argument \\structure analyzer \end{tabular}& 67\% average accuracy \\
\hline
McElvain et al. (2019b) & \begin{tabular}[c]{@{}c@{}}Legal QA system \\using pre-trained models\end{tabular} & \begin{tabular}[c]{@{}c@{}}22M documents
classified \\ to over 120K legal topics \end{tabular} & \begin{tabular}[c]{@{}c@{}}Use of pre-trained models\\ and fine-tuning on legal dataset \end{tabular}& \begin{tabular}[c]{@{}c@{}} 90\% Answered at 3 for correct answers, \\1.5\% Answered at 3 for incorrect answers \end{tabular}\\
\hline
Kim et al. (2017) & \begin{tabular}[c]{@{}c@{}}Legal QA using CNN\end{tabular} & \begin{tabular}[c]{@{}c@{}}COLIEE  \citep{kim2015coliee}\end{tabular} & \begin{tabular}[c]{@{}c@{}}Exploiting legal information \\retrieval and textual \\entailment using CNN \end{tabular}& 63.87\% \\
\hline
Kim et al. (2017) & \begin{tabular}[c]{@{}c@{}}Legal information \\retrieval and QA \end{tabular}& \begin{tabular}[c]{@{}c@{}}Japanese civil \\law articles and \\legal bar exams \end{tabular} & \begin{tabular}[c]{@{}c@{}}Combination of tf-idf \\and SVM re-ranking model,\\ lemmatization and dependency parsing \end{tabular}& \begin{tabular}[c]{@{}c@{}}62.14\% on the dry run dataset \\and 55.71\%, 55.79\% for \\Phase 2 and 3 respectively \end{tabular}\\
\hline
Duong et al. (2014) & \begin{tabular}[c]{@{}c@{}}Vietnamese QA system \end{tabular}& \begin{tabular}[c]{@{}c@{}}Vietnam’s legal documents\end{tabular}& \begin{tabular}[c]{@{}c@{}}Utilization of Vietnamese \\resources and tools, \\similarity-based model,\\ and a combination of rule-based \\and machine learning methods \end{tabular}& \begin{tabular}[c]{@{}c@{}}70\% precision \end{tabular}\\
\hline
Kim et al. (2018) & \begin{tabular}[c]{@{}c@{}}Textual entailment for \\ legal question answering	 \end{tabular}& \begin{tabular}[c]{@{}c@{}}COLIEE 2014 training data for training,\\ and the COLIEE 2015 test data for validation\end{tabular}& \begin{tabular}[c]{@{}c@{}}Development of a legal question answering \\system using Siamese CNNs, preprocessing of data by \\removing stop words and performing stemming,\\ use of a three-layer CNN to extract word features and a max \\pooling layer, use of dropout to \\prevent overfitting	 \end{tabular}& \begin{tabular}[c]{@{}c@{}}64.25\%  \end{tabular}\\

\hline
Martinez-Gil et al. (2019)  & \begin{tabular}[c]{@{}c@{}}Analyzing co-occurrence\\ patterns in unstructured text\\ corpora	\end{tabular}& \begin{tabular}[c]{@{}c@{}}Legal questions randomly \\selected from books \\from the Oxford University \\Press	\end{tabular}& \begin{tabular}[c]{@{}c@{}}A new method for the automatic \\answer of multiple choice questions \\in the legal domain, ability to \\reduce workload for professionals\\ in the legal sector, \\ability to be extrapolated to\\ other specific domains		 \end{tabular}& \begin{tabular}[c]{@{}c@{}}65\%  \end{tabular}\\

\hline
Hoshino et al. (2019)   & \begin{tabular}[c]{@{}c@{}}Predicate Argument\\ Structure Analysis	\end{tabular}& \begin{tabular}[c]{@{}c@{}}COLIEE 2018  \citep{kim2015coliee}	\end{tabular}& \begin{tabular}[c]{@{}c@{}}Created legal term dictionary\\, Synonym dictionary for predicates,\\ Person estimation feature,\\ Four types of question answering modules			 \end{tabular}& \begin{tabular}[c]{@{}c@{}}70\%  \end{tabular}\\

\hline

Alotaibi et al. (2022)    & \begin{tabular}[c]{@{}c@{}} Combination of retrieval-based \\and generative-based techniques with \\incorporation of prior knowledge sources \\such as previous questions, \\question categories, and \\Islamic Jurisprudential reference books		\end{tabular}& \begin{tabular}[c]{@{}c@{}}Custom dataset	\end{tabular}& \begin{tabular}[c]{@{}c@{}}Reduced workload on human experts \\by providing relevant and high-quality \\answers to aid in Muslims'\\ daily life decisions.	\end{tabular}& \begin{tabular}[c]{@{}c@{}} 0.60 precision, 0.40 recall,\\ 0.48 F1 and 0.037 for METEOR  \end{tabular}\\

\hline

Hoppe et al. 2021b   & \begin{tabular}[c]{@{}c@{}}Intelligent Legal Advisor	\end{tabular}& \begin{tabular}[c]{@{}c@{}}German legal documents	\end{tabular}& \begin{tabular}[c]{@{}c@{}}Semantic document retrieval \\and QA using state-of-the-art \\technologies in NLP, semantic search, \\and knowledge engineering		\end{tabular}& \begin{tabular}[c]{@{}c@{}} 0.84 Recall\\ 0.73 MAP  \end{tabular}\\

\hline
\end{tabular}
\end{adjustbox}
\caption{Comparison of legal question answering methods in terms of approach, dataset, key contributions, and accuracy.}
\label{tab:comparison}
\end{table}
\end{landscape}

\section{Discussion}
In this section, we will discuss and summarize the latest trends in legal QA processing and propose some possible extensions while also discussing freely available datasets, evaluation metrics, evaluation tools, and language resources and toolkits. We will begin by presenting various legal QA approaches and then delve deeper into the current state of the field.

To gain a better understanding of the current trends in Legal QA methods, we begin by showcasing Figure \ref{fig:Chart about Table analysis publications}, which illustrates the number of publication years. The figure reveals a steady rise in the total number of approaches since 2014. Several collaborative methods have been developed to leverage the public's efforts in improving the accuracy of legal QA systems. The latest one was published in 2022. \citep{WOS000820962300044} is a Knowledge Augmented BERT2BERT Automated Questions-Answering system for Jurisprudential Legal Opinions. It is a Question-Answering (QA) system based on retrieval augmented generative transformer model for jurisprudential legal questions. The system is designed to solve the problem of jurisprudential legal rules that govern how Muslims react and interact.

The COLIEE competitions held in 2019, 2022, and the upcoming one in 2023 have been instrumental in advancing the field of legal question answering (QA) by providing a standardized platform for evaluating submitted approaches on the same dataset, using the same metrics, and even the same published evaluation tool. The competitions have been running since 2007 and have evolved over time to include a range of subtasks related to legal information extraction and entailment. By participating in these competitions, researchers and practitioners have been able to test and refine their techniques and approaches in a standardized environment, thus paving the way for more effective and accurate legal problem-solving. For better views on the performance of methods on each dataset, we provide a summary table \ref{tab:comparison}. This table summarizes methods with respect to working or being tested on either public datasets or private ones. 
%In the section \ref{qa_method} that focused on question answering, it is common to list all the matrices that have been used in the study. 
This section is important as it provides a clear overview of the different features that have been considered in the model, and helps readers to understand the methodology and approach taken by the authors.

After thoroughly analyzing and studying several research papers in the field of Legal QA, we have identified several common themes and approaches that could be used as guidelines and potential directions for future research. In the following two sub-sections, we recommend some guidelines and potential directions in the two following sub-sections.

\subsection{Guidelines for Legal QA}

Based on our analysis of the literature, we recommend that future Legal QA research focus on the following guidelines:
\begin{itemize}
    \item Use of legal-specific knowledge bases: Utilizing legal-specific knowledge can help to improve the accuracy and efficiency of Legal QA systems.
    \item Incorporation of domain-specific features: Incorporating domain-specific features such as legal concepts, entities, and relations can improve the performance of Legal QA systems.

    \item Development of multi-stage models: Developing multi-stage models that incorporate both retrieval and extraction stages can help to improve the accuracy and efficiency of Legal QA systems.
    \item Data augmentation techniques can be used to artificially expand the size of a given dataset, which can help to improve the performance of machine learning models. In the case of Legal QA, question answer data augmentation involves generating additional training examples by modifying the phrasing or wording of existing questions and answers in the dataset. This approach can help to increase the diversity of the training data and improve the model's ability to handle variations in the wording of questions and answers.

\end{itemize}

\subsection{Potential extensions}
Along with the guidelines, we suggest some potential directions for developing post-processing approaches.
\begin{itemize}
\item While several datasets are commonly utilized to evaluate the performance of various Legal QA approaches, only a limited number of these datasets are freely accessible. These publicly available datasets serve as valuable resources, enabling researchers to compare the effectiveness of their methods and gain a better understanding of their strengths and limitations. However, it should be noted that even when using the same dataset, the manner in which the training, development, and testing data are divided can lead to challenges when attempting to make effective comparisons between different approaches. This highlights the importance of establishing clear and consistent evaluation protocols in Legal QA research, which can help to ensure that results are reproducible and comparable across studies.
\item Integration of Explainable AI techniques: One potential extension is to explore the integration of explainable AI technique techniques such as attention visualization and explanation generation. This can help to provide transparency and interpretability to Legal QA systems, enabling users to understand the reasoning behind the system's outputs.

\item Development of interactive Legal QA systems: Another potential extension is the development of interactive Legal QA systems that allow users to interact with the system and provide feedback on the accuracy and relevance of the system's outputs. This can help to improve the user experience and enable the system to learn from user feedback.

\item Investigation of Legal QA for specific legal domains: While Legal QA has been primarily focused on open-domain question answering, there is a need to investigate Legal QA for specific legal domains such as intellectual property, tax law, and criminal law. This can help to develop domain-specific Legal QA systems that are tailored to the unique requirements and challenges of each domain.

\item As the majority of existing approaches in Legal QA are tailored to English language, it is crucial to focus on the development of methods and datasets for Legal QA in other languages.
\end{itemize}

%and future work
\section{Conclusion  }
Legal Question Answering (LQA) is a rapidly growing research field that aims to develop models capable of answering legal questions automatically. The survey discusses a comprehensive review of recent research on legal  QA systems. We highlight the key contributions of these studies, including the development of new taxonomies for legal QA systems, the use of advanced  NLP techniques such as deep learning and semantic analysis, and the incorporation of abundant resources such as legal dictionaries and knowledge bases. The survey also discusses the various challenges that legal QA systems still face and potential directions for future research in this field.

\bibliographystyle{abbrvnat} 
\bibliography{elsarticle-template-num}

\begin{thebibliography}{99}
\providecommand{\natexlab}[1]{#1}
\providecommand{\url}[1]{\texttt{#1}}
\expandafter\ifx\csname urlstyle\endcsname\relax
  \providecommand{\doi}[1]{doi: #1}\else
  \providecommand{\doi}{doi: \begingroup \urlstyle{rm}\Url}\fi

\bibitem[Abbasiantaeb and Momtazi(2021)]{abbasiantaeb2021text}
Z.~Abbasiantaeb and S.~Momtazi.
\newblock Text-based question answering from information retrieval and deep
  neural network perspectives: A survey.
\newblock \emph{Wiley Interdisciplinary Reviews: Data Mining and Knowledge
  Discovery}, 11\penalty0 (6):\penalty0 e1412, 2021.

\bibitem[Abdallah and Jatowt(2023)]{abdallah2023generator}
A.~Abdallah and A.~Jatowt.
\newblock Generator-retriever-generator: A novel approach to open-domain
  question answering.
\newblock \emph{arXiv preprint arXiv:2307.11278}, 2023.

\bibitem[Abdallah et~al.(2020{\natexlab{a}})Abdallah, Hamada, and
  Nurseitov]{abdallah2020attention}
A.~Abdallah, M.~Hamada, and D.~Nurseitov.
\newblock Attention-based fully gated cnn-bgru for russian handwritten text.
\newblock \emph{Journal of Imaging}, 6\penalty0 (12):\penalty0 141,
  2020{\natexlab{a}}.

\bibitem[Abdallah et~al.(2020{\natexlab{b}})Abdallah, Kasem, Hamada, and
  Sdeek]{abdallah2020automated}
A.~Abdallah, M.~Kasem, M.~A. Hamada, and S.~Sdeek.
\newblock Automated question-answer medical model based on deep learning
  technology.
\newblock In \emph{Proceedings of the 6th International Conference on
  Engineering \& MIS 2020}, pages 1--8, 2020{\natexlab{b}}.

\bibitem[Ahmad et~al.(2020)Ahmad, Chi, Tian, and Chang]{ahmad2020policyqa}
W.~U. Ahmad, J.~Chi, Y.~Tian, and K.-W. Chang.
\newblock Policyqa: A reading comprehension dataset for privacy policies.
\newblock \emph{arXiv preprint arXiv:2010.02557}, 2020.

\bibitem[Allam and Haggag(2012)]{allam2012question}
A.~M.~N. Allam and M.~H. Haggag.
\newblock The question answering systems: A survey.
\newblock \emph{International Journal of Research and Reviews in Information
  Sciences (IJRRIS)}, 2\penalty0 (3), 2012.

\bibitem[Alotaibi et~al.(2022)Alotaibi, Munshi, Farag, Rakha, Al~Sallab, and
  Alotaibi]{WOS000820962300044}
S.~S. Alotaibi, A.~A. Munshi, A.~T. Farag, O.~E. Rakha, A.~A. Al~Sallab, and
  M.~Alotaibi.
\newblock Kab: Knowledge augmented bert2bert automated questions-answering
  system for jurisprudential legal opinions.
\newblock \emph{INTERNATIONAL JOURNAL OF COMPUTER SCIENCE AND NETWORK
  SECURITY}, 22\penalty0 (6):\penalty0 346--356, JUN 30 2022.
\newblock ISSN 1738-7906.
\newblock \doi{10.22937/IJCSNS.2022.22.6.44}.

\bibitem[Askari et~al.(2022)Askari, Verberne, and Pasi]{askari2022expert}
A.~Askari, S.~Verberne, and G.~Pasi.
\newblock Expert finding in legal community question answering.
\newblock In \emph{European Conference on Information Retrieval}, pages 22--30.
  Springer, 2022.

\bibitem[Bach et~al.(2017)Bach, Thien, Phuong, et~al.]{bach2017question}
N.~X. Bach, T.~H.~N. Thien, T.~M. Phuong, et~al.
\newblock Question analysis for vietnamese legal question answering.
\newblock In \emph{2017 9th International Conference on Knowledge and Systems
  Engineering (KSE)}, pages 154--159. IEEE, 2017.

\bibitem[Baral(2003)]{baral2003knowledge}
C.~Baral.
\newblock \emph{Knowledge representation, reasoning and declarative problem
  solving}.
\newblock Cambridge university press, 2003.

\bibitem[Bongard et~al.(2022)Bongard, Held, and Habernal]{bongard2022legal}
L.~Bongard, L.~Held, and I.~Habernal.
\newblock The legal argument reasoning task in civil procedure.
\newblock \emph{arXiv preprint arXiv:2211.02950}, 2022.

\bibitem[Brown et~al.(2020{\natexlab{a}})Brown, Mann, Ryder, Subbiah, Kaplan,
  Dhariwal, Neelakantan, Shyam, Sastry, Askell, et~al.]{brown2020language}
T.~Brown, B.~Mann, N.~Ryder, M.~Subbiah, J.~D. Kaplan, P.~Dhariwal,
  A.~Neelakantan, P.~Shyam, G.~Sastry, A.~Askell, et~al.
\newblock Language models are few-shot learners.
\newblock \emph{Advances in neural information processing systems},
  33:\penalty0 1877--1901, 2020{\natexlab{a}}.

\bibitem[Brown et~al.(2020{\natexlab{b}})Brown, Mann, Ryder, Subbiah, Kaplan,
  Dhariwal, Neelakantan, Shyam, Sastry, Askell, Agarwal, Herbert-Voss, Krueger,
  Henighan, Child, Ramesh, Ziegler, Wu, Winter, Hesse, Chen, Sigler, Litwin,
  Gray, Chess, Clark, Berner, McCandlish, Radford, Sutskever, and
  Amodei]{brown2020gpt3}
T.~B. Brown, B.~Mann, N.~Ryder, M.~Subbiah, J.~Kaplan, P.~Dhariwal,
  A.~Neelakantan, P.~Shyam, G.~Sastry, A.~Askell, S.~Agarwal, A.~Herbert-Voss,
  G.~Krueger, T.~Henighan, R.~Child, A.~Ramesh, D.~M. Ziegler, J.~Wu,
  C.~Winter, C.~Hesse, M.~Chen, E.~Sigler, M.~Litwin, S.~Gray, B.~Chess,
  J.~Clark, C.~Berner, S.~McCandlish, A.~Radford, I.~Sutskever, and D.~Amodei.
\newblock Language models are few-shot learners, 2020{\natexlab{b}}.
\newblock URL \url{https://arxiv.org/abs/2005.14165}.

\bibitem[Cadene et~al.(2019)Cadene, Dancette, Cord, Parikh,
  et~al.]{cadene2019rubi}
R.~Cadene, C.~Dancette, M.~Cord, D.~Parikh, et~al.
\newblock Rubi: Reducing unimodal biases for visual question answering.
\newblock \emph{Advances in neural information processing systems}, 32, 2019.

\bibitem[Chen et~al.(2020)Chen, Han, and Wang]{chen2020multimodal}
C.~Chen, D.~Han, and J.~Wang.
\newblock Multimodal encoder-decoder attention networks for visual question
  answering.
\newblock \emph{IEEE Access}, 8:\penalty0 35662--35671, 2020.

\bibitem[Choi et~al.(2018)Choi, He, Iyyer, Yatskar, Yih, Choi, Liang, and
  Zettlemoyer]{choi2018quac}
E.~Choi, H.~He, M.~Iyyer, M.~Yatskar, W.-t. Yih, Y.~Choi, P.~Liang, and
  L.~Zettlemoyer.
\newblock Quac: Question answering in context.
\newblock \emph{arXiv preprint arXiv:1808.07036}, 2018.

\bibitem[Clarke and Terra(2003)]{clarke2003passage}
C.~L. Clarke and E.~L. Terra.
\newblock Passage retrieval vs. document retrieval for factoid question
  answering.
\newblock In \emph{Proceedings of the 26th annual international ACM SIGIR
  conference on Research and development in informaion retrieval}, pages
  427--428, 2003.

\bibitem[Collarana et~al.(2018)Collarana, Heuss, Lehmann, Lytra, Maheshwari,
  Nedelchev, Schmidt, and Trivedi]{collarana2018question}
D.~Collarana, T.~Heuss, J.~Lehmann, I.~Lytra, G.~Maheshwari, R.~Nedelchev,
  T.~Schmidt, and P.~Trivedi.
\newblock A question answering system on regulatory documents.
\newblock In \emph{Legal Knowledge and Information Systems}, pages 41--50. IOS
  Press, 2018.

\bibitem[Devlin et~al.(2018)Devlin, Chang, Lee, and Toutanova]{devlin2018bert}
J.~Devlin, M.-W. Chang, K.~Lee, and K.~Toutanova.
\newblock Bert: Pre-training of deep bidirectional transformers for language
  understanding.
\newblock \emph{arXiv preprint arXiv:1810.04805}, 2018.

\bibitem[Do et~al.(2017)Do, Nguyen, Tran, Nguyen, and Nguyen]{do2017legal}
P.-K. Do, H.-T. Nguyen, C.-X. Tran, M.-T. Nguyen, and M.-L. Nguyen.
\newblock Legal question answering using ranking svm and deep convolutional
  neural network.
\newblock \emph{arXiv preprint arXiv:1703.05320}, 2017.

\bibitem[Duan et~al.(2019)Duan, Wang, Wang, Ma, Cui, Wu, Wang, Liu, Huo, Hu,
  et~al.]{duan2019cjrc}
X.~Duan, B.~Wang, Z.~Wang, W.~Ma, Y.~Cui, D.~Wu, S.~Wang, T.~Liu, T.~Huo,
  Z.~Hu, et~al.
\newblock Cjrc: A reliable human-annotated benchmark dataset for chinese
  judicial reading comprehension.
\newblock In \emph{China National Conference on Chinese Computational
  Linguistics}, pages 439--451. Springer, 2019.

\bibitem[Duong and Ho(2014)]{duong2014vietnamese}
H.-T. Duong and B.-Q. Ho.
\newblock A vietnamese question answering system in vietnam’s legal
  documents.
\newblock In \emph{Computer Information Systems and Industrial Management: 13th
  IFIP TC8 International Conference, CISIM 2014, Ho Chi Minh City, Vietnam,
  November 5-7, 2014. Proceedings 14}, pages 186--197. Springer, 2014.

\bibitem[Ezzeldin and Shaheen(2012)]{ezzeldin2012survey}
A.~M. Ezzeldin and M.~Shaheen.
\newblock A survey of arabic question answering: challenges, tasks, approaches,
  tools, and future trends.
\newblock In \emph{Proceedings of The 13th international Arab conference on
  information technology (ACIT 2012)}, pages 1--8, 2012.

\bibitem[Fawei et~al.(2019)Fawei, Pan, Kollingbaum, and Wyner]{fawei2019semi}
B.~Fawei, J.~Z. Pan, M.~Kollingbaum, and A.~Z. Wyner.
\newblock A semi-automated ontology construction for legal question answering.
\newblock \emph{New Generation Computing}, 37\penalty0 (4):\penalty0 453--478,
  2019.

\bibitem[Ganguly et~al.(2023)Ganguly, Conrad, Ghosh, Ghosh, Goyal,
  Bhattacharya, Nigam, and Paul]{ECIR2023}
D.~Ganguly, J.~G. Conrad, K.~Ghosh, S.~Ghosh, P.~Goyal, P.~Bhattacharya, S.~K.
  Nigam, and S.~Paul.
\newblock Legal ir and nlp: The history, challenges, and state-of-the-art.
\newblock In \emph{Advances in Information Retrieval}, pages 331--340, Cham,
  2023. Springer Nature Switzerland.

\bibitem[Glannon(2018)]{glannon2018glannon}
J.~W. Glannon.
\newblock \emph{Glannon Guide to Civil Procedure: Learning Civil Procedure
  Through Multiple-Choice Questions and Analysis}.
\newblock Aspen Publishing, 2018.

\bibitem[Golub and He(2016)]{golub2016character}
D.~Golub and X.~He.
\newblock Character-level question answering with attention.
\newblock \emph{arXiv preprint arXiv:1604.00727}, 2016.

\bibitem[Goutte and Gaussier(2005)]{inproceedings1}
C.~Goutte and E.~Gaussier.
\newblock A probabilistic interpretation of precision, recall and f-score, with
  implication for evaluation.
\newblock volume 3408, pages 345--359, 04 2005.
\newblock ISBN 978-3-540-25295-5.
\newblock \doi{10.1007/978-3-540-31865-1_25}.

\bibitem[Guda et~al.(2011)Guda, Sanampudi, and Manikyamba]{guda2011approaches}
V.~Guda, S.~K. Sanampudi, and I.~L. Manikyamba.
\newblock Approaches for question answering systems.
\newblock \emph{International Journal of Engineering science and technology
  (IJEST)}, 3\penalty0 (2):\penalty0 990--995, 2011.

\bibitem[Guo et~al.(2018)Guo, Zhou, Li, and Wang]{guo2018hierarchical}
D.~Guo, W.~Zhou, H.~Li, and M.~Wang.
\newblock Hierarchical lstm for sign language translation.
\newblock In \emph{Proceedings of the AAAI conference on artificial
  intelligence}, volume~32, 2018.

\bibitem[Gupta and Gupta(2012)]{gupta2012survey}
P.~Gupta and V.~Gupta.
\newblock A survey of text question answering techniques.
\newblock \emph{International Journal of Computer Applications}, 53\penalty0
  (4), 2012.

\bibitem[Hayes-Roth(1985)]{hayes1985rule}
F.~Hayes-Roth.
\newblock Rule-based systems.
\newblock \emph{Communications of the ACM}, 28\penalty0 (9):\penalty0 921--932,
  1985.

\bibitem[He and Golub(2016)]{he2016character}
X.~He and D.~Golub.
\newblock Character-level question answering with attention.
\newblock In \emph{Proceedings of the 2016 conference on empirical methods in
  natural language processing}, pages 1598--1607, 2016.

\bibitem[Hendrycks et~al.(2021)Hendrycks, Burns, Chen, and
  Ball]{hendrycks2021cuad}
D.~Hendrycks, C.~Burns, A.~Chen, and S.~Ball.
\newblock Cuad: An expert-annotated nlp dataset for legal contract review.
\newblock \emph{arXiv preprint arXiv:2103.06268}, 2021.

\bibitem[Hochreiter and Schmidhuber(1997)]{hochreiter1997long}
S.~Hochreiter and J.~Schmidhuber.
\newblock Long short-term memory.
\newblock \emph{Neural computation}, 9\penalty0 (8):\penalty0 1735--1780, 1997.

\bibitem[Hoppe et~al.(2021{\natexlab{a}})Hoppe, Pelkmann, Migenda, H{\"o}tte,
  and Schenck]{hoppe2021towards}
C.~Hoppe, D.~Pelkmann, N.~Migenda, D.~H{\"o}tte, and W.~Schenck.
\newblock Towards intelligent legal advisors for document retrieval and
  question-answering in german legal documents.
\newblock In \emph{2021 IEEE Fourth International Conference on Artificial
  Intelligence and Knowledge Engineering (AIKE)}, pages 29--32. IEEE,
  2021{\natexlab{a}}.

\bibitem[Hoppe et~al.(2021{\natexlab{b}})Hoppe, Pelkmann, Migenda, Hötte, and
  Schenck]{9723721}
C.~Hoppe, D.~Pelkmann, N.~Migenda, D.~Hötte, and W.~Schenck.
\newblock Towards intelligent legal advisors for document retrieval and
  question-answering in german legal documents.
\newblock In \emph{2021 IEEE Fourth International Conference on Artificial
  Intelligence and Knowledge Engineering (AIKE)}, pages 29--32,
  2021{\natexlab{b}}.
\newblock \doi{10.1109/AIKE52691.2021.00011}.

\bibitem[Hoshino et~al.(2019)Hoshino, Taniguchi, Kiyota, and
  Kano]{hoshino2019question}
R.~Hoshino, R.~Taniguchi, N.~Kiyota, and Y.~Kano.
\newblock Question answering system for legal bar examination using predicate
  argument structure.
\newblock In \emph{New Frontiers in Artificial Intelligence: JSAI-isAI 2018
  Workshops, JURISIN, AI-Biz, SKL, LENLS, IDAA, Yokohama, Japan, November
  12--14, 2018, Revised Selected Papers}, pages 207--220. Springer, 2019.

\bibitem[Huang et~al.(2020)Huang, Jiang, Qu, and Yang]{huang2020aila}
W.~Huang, J.~Jiang, Q.~Qu, and M.~Yang.
\newblock Aila: A question answering system in the legal domain.
\newblock In \emph{IJCAI}, pages 5258--5260, 2020.

\bibitem[Kadam et~al.(2015)Kadam, Joshi, Shinde, and Medhane]{kadam2015notice}
A.~D. Kadam, S.~D. Joshi, S.~V. Shinde, and S.~P. Medhane.
\newblock Notice of removal: Question answering search engine short review and
  road-map to future qa search engine.
\newblock In \emph{2015 International Conference on Electrical, Electronics,
  Signals, Communication and Optimization (EESCO)}, pages 1--8. IEEE, 2015.

\bibitem[Kassner and Sch{\"u}tze(2020)]{kassner-schutze-2020-bert}
N.~Kassner and H.~Sch{\"u}tze.
\newblock {BERT}-k{NN}: Adding a k{NN} search component to pretrained language
  models for better {QA}.
\newblock In \emph{Findings of the Association for Computational Linguistics:
  EMNLP 2020}, pages 3424--3430, Online, Nov. 2020. Association for
  Computational Linguistics.
\newblock \doi{10.18653/v1/2020.findings-emnlp.307}.
\newblock URL \url{https://aclanthology.org/2020.findings-emnlp.307}.

\bibitem[Khandelwal et~al.(2020)Khandelwal, Levy, Jurafsky, Zettlemoyer, and
  Lewis]{khandelwal20generalization}
U.~Khandelwal, O.~Levy, D.~Jurafsky, L.~Zettlemoyer, and M.~Lewis.
\newblock {Generalization through Memorization: Nearest Neighbor Language
  Models}.
\newblock In \emph{International Conference on Learning Representations
  (ICLR)}, 2020.

\bibitem[Kien et~al.(2020)Kien, Nguyen, Bach, Tran, Le~Nguyen, and
  Phuong]{kien2020answering}
P.~M. Kien, H.-T. Nguyen, N.~X. Bach, V.~Tran, M.~Le~Nguyen, and T.~M. Phuong.
\newblock Answering legal questions by learning neural attentive text
  representation.
\newblock In \emph{Proceedings of the 28th International Conference on
  Computational Linguistics}, pages 988--998, 2020.

\bibitem[Kim et~al.(2014{\natexlab{a}})Kim, Xu, and Goebel]{kim2014alberta}
M.~Kim, Y.~Xu, and R.~Goebel.
\newblock Alberta-kxg: legal question answering using ranking svm and
  syntactic/semantic similarity.
\newblock In \emph{JURISIN Workshop}, 2014{\natexlab{a}}.

\bibitem[Kim et~al.(2014{\natexlab{b}})Kim, Xu, Goebel, and
  Satoh]{kim2014answering}
M.-Y. Kim, Y.~Xu, R.~Goebel, and K.~Satoh.
\newblock Answering yes/no questions in legal bar exams.
\newblock In \emph{JSAI International Symposium on Artificial Intelligence},
  pages 199--213. Springer, 2014{\natexlab{b}}.

\bibitem[Kim et~al.(2015{\natexlab{a}})Kim, Goebel, and Ken]{kim2015coliee}
M.-Y. Kim, R.~Goebel, and S.~Ken.
\newblock Coliee-2015: evaluation of legal question answering.
\newblock In \emph{Ninth International Workshop on Juris-informatics (JURISIN
  2015)}, 2015{\natexlab{a}}.

\bibitem[Kim et~al.(2015{\natexlab{b}})Kim, Xu, and Goebel]{kim2015applying}
M.-Y. Kim, Y.~Xu, and R.~Goebel.
\newblock Applying a convolutional neural network to legal question answering.
\newblock In \emph{JSAI International Symposium on Artificial Intelligence},
  pages 282--294. Springer, 2015{\natexlab{b}}.

\bibitem[Kim et~al.(2015{\natexlab{c}})Kim, Xu, and
  Goebel]{kim2015convolutional}
M.-Y. Kim, Y.~Xu, and R.~Goebel.
\newblock A convolutional neural network in legal question answering.
\newblock In \emph{JURISIN Workshop}, 2015{\natexlab{c}}.

\bibitem[Kim et~al.(2017{\natexlab{a}})Kim, Xu, and Goebel]{kim2017applying}
M.-Y. Kim, Y.~Xu, and R.~Goebel.
\newblock Applying a convolutional neural network to legal question answering.
\newblock In \emph{New Frontiers in Artificial Intelligence: JSAI-isAI 2015
  Workshops, LENLS, JURISIN, AAA, HAT-MASH, TSDAA, ASD-HR, and SKL, Kanagawa,
  Japan, November 16-18, 2015, Revised Selected Papers}, pages 282--294.
  Springer, 2017{\natexlab{a}}.

\bibitem[Kim et~al.(2017{\natexlab{b}})Kim, Xu, Lu, and
  Goebel]{kim2017question}
M.-Y. Kim, Y.~Xu, Y.~Lu, and R.~Goebel.
\newblock Question answering of bar exams by paraphrasing and legal text
  analysis.
\newblock In \emph{New Frontiers in Artificial Intelligence: JSAI-isAI 2016
  Workshops, LENLS, HAT-MASH, AI-Biz, JURISIN and SKL, Kanagawa, Japan,
  November 14-16, 2016, Revised Selected Papers}, pages 299--313. Springer,
  2017{\natexlab{b}}.

\bibitem[Kim et~al.(2018)Kim, Lu, and Goebel]{kim2018textual}
M.-Y. Kim, Y.~Lu, and R.~Goebel.
\newblock Textual entailment in legal bar exam question answering using deep
  siamese networks.
\newblock In \emph{New Frontiers in Artificial Intelligence: JSAI-isAI
  Workshops, JURISIN, SKL, AI-Biz, LENLS, AAA, SCIDOCA, kNeXI, Tsukuba, Tokyo,
  November 13-15, 2017, Revised Selected Papers 9}, pages 35--48. Springer,
  2018.

\bibitem[Kolomiyets and Moens(2011)]{KOLOMIYETS20115412}
O.~Kolomiyets and M.-F. Moens.
\newblock A survey on question answering technology from an information
  retrieval perspective.
\newblock \emph{Information Sciences}, 181\penalty0 (24):\penalty0 5412--5434,
  2011.
\newblock ISSN 0020-0255.
\newblock \doi{https://doi.org/10.1016/j.ins.2011.07.047}.
\newblock URL
  \url{https://www.sciencedirect.com/science/article/pii/S0020025511003860}.

\bibitem[Komeili et~al.(2021)Komeili, Shuster, and Weston]{komeili2021internet}
M.~Komeili, K.~Shuster, and J.~Weston.
\newblock Internet-augmented dialogue generation.
\newblock \emph{arXiv preprint arXiv:2107.07566}, 2021.

\bibitem[Krizhevsky et~al.(2017)Krizhevsky, Sutskever, and
  Hinton]{krizhevsky2017imagenet}
A.~Krizhevsky, I.~Sutskever, and G.~E. Hinton.
\newblock Imagenet classification with deep convolutional neural networks.
\newblock \emph{Communications of the ACM}, 60\penalty0 (6):\penalty0 84--90,
  2017.

\bibitem[Lewis et~al.(2019)Lewis, Denoyer, and Riedel]{lewis2019unsupervised}
P.~Lewis, L.~Denoyer, and S.~Riedel.
\newblock Unsupervised question answering by cloze translation.
\newblock \emph{arXiv preprint arXiv:1906.04980}, 2019.

\bibitem[Lin et~al.(2017)Lin, Feng, Santos, Yu, Xiang, Zhou, and
  Bengio]{lin2017structured}
Z.~Lin, M.~Feng, C.~N.~d. Santos, M.~Yu, B.~Xiang, B.~Zhou, and Y.~Bengio.
\newblock A structured self-attentive sentence embedding.
\newblock \emph{arXiv preprint arXiv:1703.03130}, 2017.

\bibitem[Liu and Lapata(2019)]{liu2019text}
Y.~Liu and M.~Lapata.
\newblock Text summarization with pretrained encoders.
\newblock \emph{arXiv preprint arXiv:1908.08345}, 2019.

\bibitem[Louis et~al.(2021)Louis, Spanakis, and Van~Dijck]{louis2021statutory}
A.~Louis, G.~Spanakis, and G.~Van~Dijck.
\newblock A statutory article retrieval dataset in french.
\newblock \emph{arXiv preprint arXiv:2108.11792}, 2021.

\bibitem[Martinez-Gil(2021)]{martinez2021survey}
J.~Martinez-Gil.
\newblock A survey on legal question answering systems.
\newblock \emph{arXiv preprint arXiv:2110.07333}, 2021.

\bibitem[Martinez-Gil et~al.(2019)Martinez-Gil, Freudenthaler, and
  Tjoa]{martinez2019multiple}
J.~Martinez-Gil, B.~Freudenthaler, and A.~M. Tjoa.
\newblock Multiple choice question answering in the legal domain using
  reinforced co-occurrence.
\newblock In \emph{International Conference on Database and Expert Systems
  Applications}, pages 138--148. Springer, 2019.

\bibitem[McElvain et~al.(2019)McElvain, Sanchez, Matthews, Teo, Pompili, and
  Custis]{10114533311843331397}
G.~McElvain, G.~Sanchez, S.~Matthews, D.~Teo, F.~Pompili, and T.~Custis.
\newblock Westsearch plus: A non-factoid question-answering system for the
  legal domain.
\newblock In \emph{Proceedings of the 42nd International ACM SIGIR Conference
  on Research and Development in Information Retrieval}, SIGIR'19, page
  1361–1364, New York, NY, USA, 2019. Association for Computing Machinery.
\newblock ISBN 9781450361729.
\newblock \doi{10.1145/3331184.3331397}.
\newblock URL \url{https://doi.org/10.1145/3331184.3331397}.

\bibitem[Morimoto et~al.(2017)Morimoto, Kubo, Sato, Shindo, and
  Matsumoto]{morimoto2017legal}
A.~Morimoto, D.~Kubo, M.~Sato, H.~Shindo, and Y.~Matsumoto.
\newblock Legal question answering system using neural attention.
\newblock \emph{COLIEE@ ICAIL}, 2017:\penalty0 79--89, 2017.

\bibitem[Mukherjee et~al.(2022)Mukherjee, Jangra, Saha, and
  Jatowt]{mukherjee2022topic}
S.~Mukherjee, A.~Jangra, S.~Saha, and A.~Jatowt.
\newblock Topic-aware multimodal summarization.
\newblock In \emph{Findings of the Association for Computational Linguistics:
  AACL-IJCNLP 2022}, pages 387--398, 2022.

\bibitem[Nguyen et~al.(2019)Nguyen, Do, Nguyen, Do, Tjiputra, and
  Tran]{nguyen2019overcoming}
B.~D. Nguyen, T.-T. Do, B.~X. Nguyen, T.~Do, E.~Tjiputra, and Q.~D. Tran.
\newblock Overcoming data limitation in medical visual question answering.
\newblock In \emph{International Conference on Medical Image Computing and
  Computer-Assisted Intervention}, pages 522--530. Springer, 2019.

\bibitem[Nie et~al.(2017)Nie, Han, Huang, Jiao, and Li]{nie2017attention}
Y.-p. Nie, Y.~Han, J.-m. Huang, B.~Jiao, and A.-p. Li.
\newblock Attention-based encoder-decoder model for answer selection in
  question answering.
\newblock \emph{Frontiers of Information Technology \& Electronic Engineering},
  18\penalty0 (4):\penalty0 535--544, 2017.

\bibitem[Pal et~al.(2012)Pal, Harper, and Konstan]{pal2012exploring}
A.~Pal, F.~M. Harper, and J.~A. Konstan.
\newblock Exploring question selection bias to identify experts and potential
  experts in community question answering.
\newblock \emph{ACM Transactions on Information Systems (TOIS)}, 30\penalty0
  (2):\penalty0 1--28, 2012.

\bibitem[Papineni et~al.(2002)Papineni, Roukos, Ward, and
  Zhu]{papineni-etal-2002-bleu}
K.~Papineni, S.~Roukos, T.~Ward, and W.-J. Zhu.
\newblock {B}leu: a method for automatic evaluation of machine translation.
\newblock In \emph{Proceedings of the 40th Annual Meeting of the Association
  for Computational Linguistics}, pages 311--318, Philadelphia, Pennsylvania,
  USA, July 2002. Association for Computational Linguistics.
\newblock \doi{10.3115/1073083.1073135}.
\newblock URL \url{https://aclanthology.org/P02-1040}.

\bibitem[Perez et~al.(2020)Perez, Lewis, Yih, Cho, and
  Kiela]{perez2020unsupervised}
E.~Perez, P.~Lewis, W.-t. Yih, K.~Cho, and D.~Kiela.
\newblock Unsupervised question decomposition for question answering.
\newblock \emph{arXiv preprint arXiv:2002.09758}, 2020.

\bibitem[Pouyanfar et~al.(2018)Pouyanfar, Sadiq, Yan, Tian, Tao, Reyes, Shyu,
  Chen, and Iyengar]{1011453234150}
S.~Pouyanfar, S.~Sadiq, Y.~Yan, H.~Tian, Y.~Tao, M.~P. Reyes, M.-L. Shyu, S.-C.
  Chen, and S.~S. Iyengar.
\newblock A survey on deep learning: Algorithms, techniques, and applications.
\newblock \emph{ACM Comput. Surv.}, 51\penalty0 (5), sep 2018.
\newblock ISSN 0360-0300.
\newblock \doi{10.1145/3234150}.
\newblock URL \url{https://doi.org/10.1145/3234150}.

\bibitem[Qu et~al.(2019)Qu, Yang, Qiu, Croft, Zhang, and Iyyer]{qu2019bert}
C.~Qu, L.~Yang, M.~Qiu, W.~B. Croft, Y.~Zhang, and M.~Iyyer.
\newblock Bert with history answer embedding for conversational question
  answering.
\newblock In \emph{Proceedings of the 42nd international ACM SIGIR conference
  on research and development in information retrieval}, pages 1133--1136,
  2019.

\bibitem[Radford et~al.(2019)Radford, Wu, Child, Luan, Amodei, and
  Sutskever]{radford2019language}
A.~Radford, J.~Wu, R.~Child, D.~Luan, D.~Amodei, and I.~Sutskever.
\newblock Language models are unsupervised multitask learners.
\newblock 2019.

\bibitem[Ravichander et~al.(2019)Ravichander, Black, Wilson, Norton, and
  Sadeh]{ravichander-etal-2019-question}
A.~Ravichander, A.~W. Black, S.~Wilson, T.~Norton, and N.~Sadeh.
\newblock Question answering for privacy policies: Combining computational and
  legal perspectives.
\newblock In \emph{Proceedings of the 2019 Conference on Empirical Methods in
  Natural Language Processing and the 9th International Joint Conference on
  Natural Language Processing (EMNLP-IJCNLP)}, pages 4949--4959, Hong Kong,
  China, Nov. 2019. Association for Computational Linguistics.
\newblock \doi{10.18653/v1/D19-1500}.
\newblock URL \url{https://www.aclweb.org/anthology/D19-1500}.

\bibitem[Reddy et~al.(2019)Reddy, Chen, and Manning]{reddy2019coqa}
S.~Reddy, D.~Chen, and C.~D. Manning.
\newblock Coqa: A conversational question answering challenge.
\newblock \emph{Transactions of the Association for Computational Linguistics},
  7:\penalty0 249--266, 2019.

\bibitem[Rumelhart et~al.(1985)Rumelhart, Hinton, and
  Williams]{rumelhart1985learning}
D.~E. Rumelhart, G.~E. Hinton, and R.~J. Williams.
\newblock Learning internal representations by error propagation.
\newblock Technical report, California Univ San Diego La Jolla Inst for
  Cognitive Science, 1985.

\bibitem[Salchner and Jatowt(2022)]{salchner-jatowt-2022-survey}
M.~F. Salchner and A.~Jatowt.
\newblock A survey of automatic text summarization using graph neural networks.
\newblock In \emph{Proceedings of the 29th International Conference on
  Computational Linguistics}, pages 6139--6150, Gyeongju, Republic of Korea,
  Oct. 2022. International Committee on Computational Linguistics.
\newblock URL \url{https://aclanthology.org/2022.coling-1.536}.

\bibitem[Sarrouti et~al.(2015)Sarrouti, Lachkar, and
  Ouatik]{sarrouti2015biomedical}
M.~Sarrouti, A.~Lachkar, and S.~E.~A. Ouatik.
\newblock Biomedical question types classification using syntactic and rule
  based approach.
\newblock In \emph{2015 7th International Joint Conference on Knowledge
  Discovery, Knowledge Engineering and Knowledge Management (IC3K)}, volume~1,
  pages 265--272. IEEE, 2015.

\bibitem[Saxe et~al.(2021)Saxe, Nelli, and Summerfield]{saxe2021if}
A.~Saxe, S.~Nelli, and C.~Summerfield.
\newblock If deep learning is the answer, what is the question?
\newblock \emph{Nature Reviews Neuroscience}, 22\penalty0 (1):\penalty0 55--67,
  2021.

\bibitem[Smith and Kandel(2018)]{smith2018verification}
S.~Smith and A.~Kandel.
\newblock \emph{Verification and validation of rule-based expert systems}.
\newblock CRC Press, 2018.

\bibitem[Song et~al.(2019)Song, Huang, and Ruan]{song2019abstractive}
S.~Song, H.~Huang, and T.~Ruan.
\newblock Abstractive text summarization using lstm-cnn based deep learning.
\newblock \emph{Multimedia Tools and Applications}, 78\penalty0 (1):\penalty0
  857--875, 2019.

\bibitem[Sovrano et~al.(2020)Sovrano, Palmirani, and Vitali]{sovrano2020legal}
F.~Sovrano, M.~Palmirani, and F.~Vitali.
\newblock Legal knowledge extraction for knowledge graph based
  question-answering.
\newblock In \emph{Legal Knowledge and Information Systems}, pages 143--153.
  IOS Press, 2020.

\bibitem[Sovrano et~al.(2021)Sovrano, Palmirani, Distefano, Sapienza, and
  Vitali]{sovrano2021dataset}
F.~Sovrano, M.~Palmirani, B.~Distefano, S.~Sapienza, and F.~Vitali.
\newblock A dataset for evaluating legal question answering on private
  international law.
\newblock In \emph{Proceedings of the Eighteenth International Conference on
  Artificial Intelligence and Law}, pages 230--234, 2021.

\bibitem[Sun et~al.(2021)Sun, Zheng, Hao, and Qiu]{sun2021nsp}
Y.~Sun, Y.~Zheng, C.~Hao, and H.~Qiu.
\newblock Nsp-bert: A prompt-based zero-shot learner through an original
  pre-training task--next sentence prediction.
\newblock \emph{arXiv preprint arXiv:2109.03564}, 2021.

\bibitem[Talmor et~al.(2018)Talmor, Herzig, Lourie, and
  Berant]{talmor2018commonsenseqa}
A.~Talmor, J.~Herzig, N.~Lourie, and J.~Berant.
\newblock Commonsenseqa: A question answering challenge targeting commonsense
  knowledge.
\newblock \emph{arXiv preprint arXiv:1811.00937}, 2018.

\bibitem[Taniguchi and Kano(2016)]{taniguchi2016legal}
R.~Taniguchi and Y.~Kano.
\newblock Legal yes/no question answering system using case-role analysis.
\newblock In \emph{JSAI International Symposium on Artificial Intelligence},
  pages 284--298. Springer, 2016.

\bibitem[Taniguchi et~al.(2019)Taniguchi, Hoshino, and
  Kano]{taniguchi2019legal}
R.~Taniguchi, R.~Hoshino, and Y.~Kano.
\newblock Legal question answering system using framenet.
\newblock In \emph{New Frontiers in Artificial Intelligence: JSAI-isAI 2018
  Workshops, JURISIN, AI-Biz, SKL, LENLS, IDAA, Yokohama, Japan, November
  12--14, 2018, Revised Selected Papers}, pages 193--206. Springer, 2019.

\bibitem[Van~Aken et~al.(2019)Van~Aken, Winter, L{\"o}ser, and
  Gers]{van2019does}
B.~Van~Aken, B.~Winter, A.~L{\"o}ser, and F.~A. Gers.
\newblock How does bert answer questions? a layer-wise analysis of transformer
  representations.
\newblock In \emph{Proceedings of the 28th ACM International Conference on
  Information and Knowledge Management}, pages 1823--1832, 2019.

\bibitem[Voorhees and Tice(2000)]{1145_Voorhees}
E.~M. Voorhees and D.~M. Tice.
\newblock Building a question answering test collection.
\newblock In \emph{Proceedings of the 23rd Annual International ACM SIGIR
  Conference on Research and Development in Information Retrieval}, SIGIR '00,
  page 200–207, New York, NY, USA, 2000. Association for Computing Machinery.
\newblock ISBN 1581132263.
\newblock \doi{10.1145/345508.345577}.
\newblock URL \url{https://doi.org/10.1145/345508.345577}.

\bibitem[Voorhees et~al.(2005)Voorhees, Harman, et~al.]{voorhees2005trec}
E.~M. Voorhees, D.~K. Harman, et~al.
\newblock \emph{TREC: Experiment and evaluation in information retrieval},
  volume~63.
\newblock Citeseer, 2005.

\bibitem[Wang et~al.(2019)Wang, Ng, Ma, Nallapati, and Xiang]{wang2019multi}
Z.~Wang, P.~Ng, X.~Ma, R.~Nallapati, and B.~Xiang.
\newblock Multi-passage bert: A globally normalized bert model for open-domain
  question answering.
\newblock \emph{arXiv preprint arXiv:1908.08167}, 2019.

\bibitem[Weston et~al.(2015)Weston, Bordes, Chopra, Rush, Van~Merri{\"e}nboer,
  Joulin, and Mikolov]{weston2015towards}
J.~Weston, A.~Bordes, S.~Chopra, A.~M. Rush, B.~Van~Merri{\"e}nboer, A.~Joulin,
  and T.~Mikolov.
\newblock Towards ai-complete question answering: A set of prerequisite toy
  tasks.
\newblock \emph{arXiv preprint arXiv:1502.05698}, 2015.

\bibitem[Xie et~al.(2020)Xie, Lu, Lin, and Lin]{xie2020faq}
R.~Xie, Y.~Lu, F.~Lin, and L.~Lin.
\newblock Faq-based question answering via knowledge anchors.
\newblock In \emph{CCF International Conference on Natural Language Processing
  and Chinese Computing}, pages 3--15. Springer, 2020.

\bibitem[Yang et~al.(2019)Yang, Hu, Qiu, Qu, Gao, Croft, Liu, Shen, and
  Liu]{yang2019hybrid}
L.~Yang, J.~Hu, M.~Qiu, C.~Qu, J.~Gao, W.~B. Croft, X.~Liu, Y.~Shen, and
  J.~Liu.
\newblock A hybrid retrieval-generation neural conversation model.
\newblock In \emph{Proceedings of the 28th ACM international conference on
  information and knowledge management}, pages 1341--1350, 2019.

\bibitem[Yin et~al.(2014)Yin, Zhang, Liu, Zhang, Xing, and
  Chen]{yin2014healthqa}
Y.~Yin, Y.~Zhang, X.~Liu, Y.~Zhang, C.~Xing, and H.~Chen.
\newblock Healthqa: A chinese qa summary system for smart health.
\newblock In \emph{International Conference on Smart Health}, pages 51--62.
  Springer, 2014.

\bibitem[Zamani et~al.(2022)Zamani, Diaz, Dehghani, Metzler, and
  Bendersky]{34774953531722}
H.~Zamani, F.~Diaz, M.~Dehghani, D.~Metzler, and M.~Bendersky.
\newblock Retrieval-enhanced machine learning.
\newblock In \emph{Proceedings of the 45th International ACM SIGIR Conference
  on Research and Development in Information Retrieval}, SIGIR '22, page
  2875–2886, New York, NY, USA, 2022. Association for Computing Machinery.
\newblock ISBN 9781450387323.
\newblock \doi{10.1145/3477495.3531722}.
\newblock URL \url{https://doi.org/10.1145/3477495.3531722}.

\bibitem[Zhang et~al.(2019)Zhang, Kishore, Wu, Weinberger, and
  Artzi]{zhang2019bertscore}
T.~Zhang, V.~Kishore, F.~Wu, K.~Q. Weinberger, and Y.~Artzi.
\newblock Bertscore: Evaluating text generation with bert.
\newblock \emph{arXiv preprint arXiv:1904.09675}, 2019.

\bibitem[Zhong et~al.(2020)Zhong, Xiao, Tu, Zhang, Liu, and Sun]{zhong2019jec}
H.~Zhong, C.~Xiao, C.~Tu, T.~Zhang, Z.~Liu, and M.~Sun.
\newblock Jec-qa: A legal-domain question answering dataset.
\newblock In \emph{Proceedings of AAAI}, 2020.

\bibitem[Zhu et~al.(2021)Zhu, Lei, Wang, Zheng, Poria, and
  Chua]{zhu2021retrieving}
F.~Zhu, W.~Lei, C.~Wang, J.~Zheng, S.~Poria, and T.-S. Chua.
\newblock Retrieving and reading: A comprehensive survey on open-domain
  question answering.
\newblock \emph{arXiv preprint arXiv:2101.00774}, 2021.

\bibitem[Zhu et~al.(2020)Zhu, Xia, Wu, He, Qin, Zhou, Li, and
  Liu]{zhu2020incorporating}
J.~Zhu, Y.~Xia, L.~Wu, D.~He, T.~Qin, W.~Zhou, H.~Li, and T.-Y. Liu.
\newblock Incorporating bert into neural machine translation.
\newblock \emph{arXiv preprint arXiv:2002.06823}, 2020.

\bibitem[Zhu et~al.(2015)Zhu, Kiros, Zemel, Salakhutdinov, Urtasun, Torralba,
  and Fidler]{zhu2015aligning}
Y.~Zhu, R.~Kiros, R.~Zemel, R.~Salakhutdinov, R.~Urtasun, A.~Torralba, and
  S.~Fidler.
\newblock Aligning books and movies: Towards story-like visual explanations by
  watching movies and reading books.
\newblock In \emph{Proceedings of the IEEE international conference on computer
  vision}, pages 19--27, 2015.

\end{thebibliography}





% \end{thebibliography}
\end{document}